\documentclass[11pt]{article}
\usepackage[]{ACL2023}

\usepackage{fancyhdr}
\fancypagestyle{plain}{%
  \fancyhf{}%
  \fancyhead[C]{Accepted to EMNLP 2025 Main Conference (Camera-Ready Version)}%
  \fancyfoot[C]{\thepage}%
}

\fancypagestyle{fancydefault}{%
  \fancyhf{}%
  \fancyhead[C]{Accepted to EMNLP 2025 Main Conference (Camera-Ready Version)}%
  \fancyfoot[C]{\thepage}%
}
\usepackage{times,latexsym}
\usepackage{xurl}
\usepackage[T1]{fontenc}
\usepackage{makecell}
\usepackage{soul}
\usepackage{graphicx}
\usepackage{xcolor}
\usepackage{booktabs}
\usepackage{microtype}
\usepackage{natbib}
\usepackage{inconsolata}
\usepackage{tabularx}
\usepackage{verbatim}
\usepackage{float}
\usepackage{bibunits}
\usepackage[utf8]{inputenc}
\usepackage{pifont}
\usepackage[normalem]{ulem}
\usepackage{amsmath}
\usepackage{amssymb}
\usepackage{colortbl}
\usepackage{longtable} 
\usepackage{caption}
\usepackage{hyperref}
\usepackage{tikz} 
\usepackage{enumitem}
\usepackage{multirow}
\usepackage[most]{tcolorbox}

\usepackage{caption}
\usepackage{array,tabularx,booktabs,siunitx}
\newcolumntype{L}[1]{>{\raggedright\arraybackslash}p{#1}}

\usepackage{array} 
\newcolumntype{N}{>{\centering\arraybackslash}p{0.7cm}} 

\DeclareCaptionFont{tenpt}{\fontsize{10pt}{12pt}\selectfont}
\newcommand{\cmark}{\ding{51}}
\newcommand{\xmark}{\ding{55}}

\usepackage{footmisc}

\definecolor{orange}{named}{black} 
\definecolor{blue}{named}{black} 

\title{{\color{blue}A Position Paper on the Automatic Generation of Machine Learning Leaderboards}}

\author{
Roelien C. Timmer$^{\clubsuit}$  \quad Yufang Hou$^{\diamondsuit}$ $^{\spadesuit}$ \quad Stephen Wan$^{\clubsuit}$\\
$^{\clubsuit}$ CSIRO Data61, Australia \\
$^{\diamondsuit}$ IT:U Interdisciplinary Transformation University Austria, Austria \\
$^{\spadesuit}$ IBM Research Europe - Ireland \\
\texttt{\{roelien.timmer, stephen.wan\}@data61.csiro.au} \\
\texttt{yufang.hou@it-u.at}
}

\begin{document}

\maketitle

\thispagestyle{plain}

\begin{abstract}
{\color{orange}An important task in machine learning (ML) research is comparing prior work, which is often performed via \emph{ML leaderboards}: a tabular overview of experiments with comparable conditions (e.g., same task, dataset, and metric).}
However, the growing volume of literature creates challenges in creating and maintaining these leaderboards. 
{\color{orange}To ease this burden, researchers have developed methods to extract \emph{leaderboard entries} from research papers for automated leaderboard curation. Yet, prior work varies in problem framing, complicating comparisons and limiting real-world applicability.}
{\color{blue}In this position paper, we present the \textbf{first overview of Automatic Leaderboard Generation (ALG) research}, identifying fundamental differences in assumptions, scope, and output formats. We propose an \textbf{ALG unified conceptual framework} to standardise how the ALG task is defined. We offer \textbf{ALG benchmarking guidelines}, including recommendations for datasets and metrics that promote fair, reproducible evaluation. Lastly, we outline \textbf{challenges and new directions for ALG}, such as, advocating for broader coverage by including all reported results and richer metadata.}

\end{abstract}

\section{Introduction}

{\color{orange}In today's fast-paced Machine Learning (ML) research environment, keeping abreast of advancements is more crucial than ever. The exponential growth in publications, exemplified by nearly a quarter of a million arXiv submissions in 2024, underscores the expanding global community of scholars and the accelerating pace of research~\cite{arxiv2024submissions}. This vast increase in information presents researchers with both rich opportunities for discovery but also makes it increasingly difficult to stay up to date.}

{\color{orange}
A key task for researchers is comparing past study outcomes to identify state-of-the-art results or benchmark against prior work. In ML, this is typically done using leaderboards: tables of experimental results under comparable conditions (e.g., \textit{task, dataset, metric}). The popularity of platforms like Papers with Code\footnote{\url{https://paperswithcode.com}} underscores their value in providing accessible, up-to-date comparisons that help researchers track progress and identify leading methods.}

\begin{figure}
    \centering
    \includegraphics[width=\linewidth]{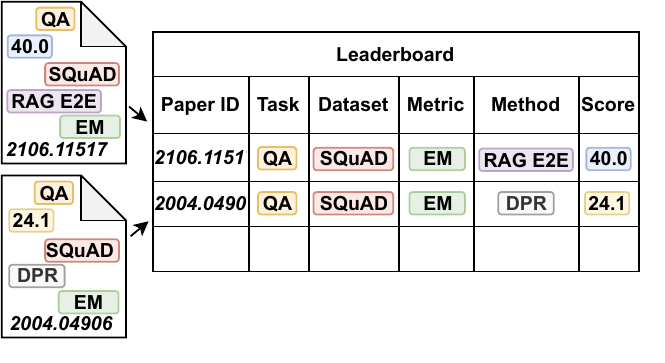}
    \caption{\fontsize{10pt}{12pt}\selectfont An example of extracting $\langle$task, dataset, metric, method, score$\rangle$ tuples from research papers to build a leaderboard\footnotemark.
    }
    \label{fig:paper_to_leaderboard}
\end{figure}
\footnotetext{An example of two LEGOBench~\cite{Singh2024LEGOBENCH} leaderboard entries summarising \newcite{siriwardhana2021fine} and \newcite{karpukhin2020dense}.} 

{\color{orange}However, leaderboards on these platforms are often incomplete or missing for certain tasks, and they typically rely on manual updates. To reduce this manual effort, recent work has focused on automatically extracting experimental outcomes (referred to here as ``tuples'') from research papers to populate leaderboards. }We refer to this body of work as \textit{Automatic Leaderboard Generation} (ALG){\color{blue}: ``A systematic process for extracting relevant experimental findings from scientific publications to create and maintain a leaderboard.''}\footnote{All acronyms used in this paper are listed in Appendix~\ref{app:acronyms} Table~\ref{tab:acronyms}.} Figure~\ref{fig:paper_to_leaderboard} illustrates an example of this process, showing the extraction of \emph{$\langle$task, dataset, metric, method, score$\rangle$} tuples from two research papers to construct a leaderboard.

Research on ALG using natural language processing (NLP) methodologies has seen significant developments in recent years. Indeed, there are still many open research questions as exemplified by the 2024 shared task on ALG~\cite{d2024overview}, underscoring the ongoing relevance of ALG. {\color{orange}This growing body of work has led to varied problem formulations and evaluation approaches, including differing assumptions about prior knowledge (\autoref{sec:reliance_on_knowledge}) and extraction scope (\autoref{sec:scope_of_extraction}), which makes comparisons across work difficult.}

{\color{blue}
This position paper makes four important contributions. First, we provide the \textbf{first overview of ALG efforts} (\autoref{sec:assumptions_preliniaries}-\autoref{sec:metrics}). By comparing prior studies side-by-side, we identify key divergences, such as variations in the assumed input scope (e.g., open vs. closed-domain) and captured results information, that previously hindered apples-to-apples comparisons. Our analysis provides a much-needed baseline map of the field, clarifying the field’s current state and identifying critical gaps.
    
Second, based on this comparison, we propose an \textbf{ALG unified conceptual framework} (\autoref{sec:methods}), essentially a problem formulation with unified terminology. This framework consolidates prior formulations into a coherent schema, providing a common language for researchers and enabling direct comparison of approaches.

Third, we provide \textbf{ALG benchmarking guidelines} (\autoref{sec:standardised_alg_benchmarking_guidelines}), to unify evaluation practices, addressing the previous lack of consensus. These guidelines establish shared standards for consistent, transparent evaluation and reliable progress tracking.

Fourth, we outline \textbf{challenges and new directions for ALG} (\autoref{sec:reflection}). We advocate expanding the extraction schema beyond just “best scores” to include all reported results (e.g., baselines, ablations) and enriching tuples with metadata (e.g., model architecture, hyperparameters) to enable more flexible result filtering.}

{\color{blue}
Ultimately, the goal of this position paper is to resolve long-standing fragmentation, establish shared standards, and open new horizons for ALG.
}

\section{{\color{blue}Overview of Problem Definition}}
\label{sec:assumptions_preliniaries}

\setlength{\tabcolsep}{3pt}
\begin{table*}[t]
\centering
\small
\begin{tabularx}{0.9\linewidth}{llcX}
\toprule

\textbf{Methodology}  &\textbf{Domain} &\textbf{Structured Data} & \textbf{Scope of Extraction} \\
\midrule
TDMS-IE~\citet{Hou2019Leaderboards} & closed  &Y&$\langle$task, dataset, metric$\rangle$ \& best score\\
PI Graph~\citet{singh2019automated} &open &Y&undefined\\
AxCell~\citet{Kardas2020AXCELL}
 & closed &Y&$\langle$task, dataset, metric$\rangle$ \& best score \\
SciREX-IE~\citet{Jain2020SCIREX} & open &Y&$\langle$task, dataset, metric, method$\rangle$, no score\\
ORKG-TDM~\citet{Kabongo2021Mining} & closed &Y&$\langle$task, dataset, metric$\rangle$, no score\\
TELIN~\citet{Yang2022TELIN} & open  &Y& $\langle$task, dataset, metric$\rangle$, best score\textsuperscript{*}\\
ORKG-LB~\citet{Kabongo2023ORKG} & closed &Y&$\langle$task, dataset, metric$\rangle$, no score\\
TDMS-PR~\citet{kabongo2024effective} & open &Y&$\langle$task, dataset, metric$\rangle$ \& best score\\
MS-PR~\citet{Singh2024LEGOBENCH} & open  &N&$\langle$task$\rangle$ \& best score\\
TDMR-PR~\citet{sahinuc2024efficient} & open  & N&$\langle$task, dataset, metric$\rangle$ \&  best score\\
\bottomrule
\multicolumn{4}{l}{\textsuperscript{*} The scope of extraction is ambiguous~\cite{Yang2022TELIN}. A response from the authors is pending for clarification.}
\end{tabularx}
\vspace{-2pt}
\caption{
\fontsize{10pt}{12pt}\selectfont {\color{blue}Characterisation of problem framing per method. Domain: open if extraction does not rely on prior knowledge, closed if restricted to a defined scope. Structured Data: Y if leaderboard tuples must appear in specific paper sections (e.g. tables or results), N otherwise. Scope of Extraction: extent of tuples extracted.}
}
\label{tab:preliminaries}
\end{table*}\vspace{-2pt}

The ALG field has seen many advances over the years.  At a broad level, the ALG task is an information extraction task, to extract a tuple containing key details of an ML experimental result.\footnote{
We acknowledge that ALG work rests on a long history of work in information extraction (IE) in scientific literature.  The full body of IE work is out of scope for this analysis but is introduced briefly in Appendix \ref{sec:related_work}.}

\citet{Hou2019Leaderboards} and \citet{singh2019automated} laid the foundation by introducing methods for extracting leaderboard tuples directly from research papers. These methodologies have since been refined and expanded upon by new methods such as AxCell~\cite{Kardas2020AXCELL}, which was put into production by Papers with Code. The most recent methodologies use prompting of pre-trained Large Language Models (LLMs), e.g. prompting Llama 2 7B ~\cite{touvron2023llama} and Mistral 7B ~\cite{jiang2023mistral} to extract $\langle$task, dataset, metric, score$\rangle$ tuples from research papers~\cite{kabongo2024effective}. 

{\color{orange}A key issue in the field is the variation in input and output expectations across studies. Table~\ref{tab:preliminaries} lists key ALG papers we examined, focusing on recent work using transformer models that enable data scaling.}\footnote{Details on prior work are in Appendix \ref{App:methodology_details}.}

We can characterise the key differences in the problem definition as concerning expectations about input and output data.  Specifically, we discuss: (1) reliance on domain knowledge, and (2) limited scope of extraction.\footnote{We also note that various works have differed in expectations on the data format (e.g., PDF or \LaTeX).  However, we do not see this as critical in hindering comparisons of results.}

\subsection{Reliance on Domain Knowledge}
\label{sec:reliance_on_knowledge}

{\color{orange}We observe that the ALG domains can be categorised as having different levels of reliance on prior domain knowledge, which ultimately impacts \textit{what} information can be extracted.}  Essentially, two variants of the problem have been previously tackled: {\textit{closed domain} and \textit{open domain}}.\footnote{The ``open domain'' category includes hybrid cases that start with no domain knowledge and incrementally builds up knowledge as publications are processed.}

\paragraph{Closed Domain:} The closed-domain ALG problem stipulates that all the entities or tuples are predefined.\footnote{As in, bound by the closed world assumption.}  In the field, there have been two subvariants that we name: (1) \textit{predefined typed entities} (PTE) and (2) \textit{predefined typed tuples} (PTT).\footnote{We borrow ``predefined'' from \citet{sahinuc2024efficient}.} 

{\color{blue}We define the \textit{predefined typed entities} (PTE) as: ``A closed-domain problem for ALG, in which the system is supplied with a finite catalogue of scientific concept classes (for instance,  specific tasks, datasets, or metrics), and extractions are confined to items from that predefined list.''} 
{\color{orange}The system may be given a declarative resource specifying entities, such as in \citet{Kardas2020AXCELL}. This could take the form of a taxonomy, a hierarchical structure of scientific concepts (e.g. tasks, datasets, metrics), or a simpler list of scientific named entities.} 

PTT is a further restriction beyond PTE in that only prescribed combinations of these science concepts are considered for establishing new tuples. {\color{blue}We define PTT as ``a closed-domain problem for ALG, in which a system is only allowed to detect leaderboard entries composed of specific, predefined combinations of known scientific concepts rather than forming any new combination.''}

In PTT variants of ALG, only predetermined combinations (often observed combinations) are used for creating new tuples (e.g., as in \citet{Hou2019Leaderboards}).

\paragraph{Open Domain:} {\color{orange}An open-domain problem allows extraction of novel entities or tuples without relying on prior knowledge (e.g. taxonomies or lists), making it less constrained. This setup is often more application-friendly, as the extraction scope is guided solely by the user's information needs.}

{\color{orange}While more appealing to users, the open-domain variant requires handling duplicates, as the same concept may appear in different forms (e.g. "ROUGE" vs. "RGE"~\cite{Jain2020SCIREX, sahinuc2024efficient}). This makes evaluation harder than in the closed domain, where canonical representations (e.g. predefined strings) enable direct accuracy measurement. Open-domain outputs may require fuzzy or semantic comparison metrics to handle variation.
}

\subsection{Scope of Extraction}\label{sec:scope_of_extraction}

{\color{orange}
Beyond differences in domain knowledge, extraction scope also varies. Prior work differs in which classes of scientific concepts, typically methodological attributes like task, dataset, method, metric, and score, are included.}

Furthermore, most work focuses only on extracting the top results from each paper, restricting each paper to a single entry per leaderboard~\cite{Hou2019Leaderboards, Kardas2020AXCELL, Hou2021TDMSci,Yang2022TELIN}. {\color{orange}If a publication presents two methods, only the top-performing one typically appears on the leaderboard. This can lead to an incomplete and potentially biased view, omitting valuable contributions such as negative results.}\footnote{{\color{blue}E.g., one may wish to compare neural networks with other machine learning methods (e.g., logistic regression, random forests) to evaluate the cost-benefit trade-off.}}
\section{{\color{blue}Overview of ALG Datasets} }
\label{sec:datasets}

\begin{table*}[ht]
\centering
\small
\begin{tabularx}{.9\linewidth}{llllllllllllllllllllll}
\toprule
&&&\multicolumn{5}{c}{\textbf{Entities}}&\multicolumn{2}{c}{\textbf{Format}}&\multicolumn{3}{c}{\textbf{Annotations}}&\textbf{Unk.}\\

\cmidrule{4-8} \cmidrule{9-10} \cmidrule{11-13} 

\textbf{Dataset}&\textbf{First Reported In}&\textbf{Variants}&\textbf{T}&\textbf{D}&\textbf{M}&\textbf{S}&\textbf{Md}&\textbf{PDF}&\textbf{\LaTeX}&\textbf{HA}&\textbf{PwC}&\textbf{NLPP}&\textbf{Ann.}\\
\toprule
ORKG-PwC&\citet{Kabongo2021Mining}& v1-v7 &\cmark&\cmark&\cmark&\xmark&\xmark&$\square$&$\square$&\xmark&\cmark&\xmark&$\square$\\
NLP-TDMS&\citet{Hou2019Leaderboards}&v1-v3&\cmark&\cmark&\cmark&\cmark&\xmark&$\square$&$\square$&\xmark&\xmark&\cmark&$\square$\\
PwC-LB&\citet{Kardas2020AXCELL}&v1-v2&\cmark&\cmark&\cmark&\cmark&\xmark&$\square$&$\square$&\xmark&\cmark&\xmark&\xmark\\
SciREX&\citet{Jain2020SCIREX}&-&\cmark&\cmark&\cmark&\xmark&\cmark&$\sim$&$\sim$&\cmark&\cmark&\xmark&\xmark\\
TDMS-Ctx&\citet{kabongo2024effective}&v1-v6&\cmark&\cmark&\cmark&\cmark&\xmark&\cmark&\xmark&\cmark&\xmark&\cmark&\cmark\\
LEGOBench&\citet{Singh2024LEGOBENCH}&-&\cmark&\cmark&\cmark&\cmark&\cmark&\cmark&\xmark&\cmark&\xmark&\cmark&\cmark\\
SciLead&\citet{sahinuc2024efficient}&-&\cmark&\cmark&\cmark&\cmark&\xmark&\cmark&\xmark&\cmark&\xmark&\xmark&\xmark\\

\bottomrule
\end{tabularx}
\caption{\fontsize{10pt}{12pt}\selectfont {\color{orange} Summary of datasets, detailing dataset \textbf{Variants}, \textbf{Entities} captured (\textbf{T} = Task, \textbf{D} = Dataset, \textbf{M} = Metric, \textbf{S} = Score, \textbf{Md} = Method), \textbf{Format} (\textbf{PDF}, \textbf{\LaTeX}), \textbf{Annotations} (\textbf{HA} = Human Annotation, \textbf{PwC} = Papers with Code, \textbf{NLPP} = NLP Progress), inclusion of unknown annotations (\textbf{Unk. Ann.}). \cmark = yes, \xmark = no, $\square$ = depends on variant, $\sim$ = use \LaTeX \hspace{0.1em} source if available, otherswise use PDF.}}\label{tab:datasets}
\end{table*}

With the growth of the field, several datasets have been proposed to evaluate ALG methods, making it hard for researchers to identify which datasets are best suited for benchmarking. {\color{orange}To guide dataset selection, Table~\ref{tab:datasets} summarises their key characteristics.\footnote{A more detailed version of this table can be found in Appendix~\ref{sec:dataset_details} Table~\ref{tab:datasets_extended}} We highlight the main dimensions along which datasets differ.} {\color{blue}The main takeaway from this table is the diversity of the datasets that have been used in past research, making it hard to make fair comparisons. We discuss the variations below.} {\color{orange}A few recent datasets offer valuable attributes: LEGOBench~\cite{Singh2024LEGOBENCH} is the largest and covers the broadest tuple scope (including score), while SciLead~\cite{sahinuc2024efficient} stands out for its exhaustive manual annotations.}

\subsection{{\color{blue}ML Experiment Science Entities}}

{\color{orange}As prior work has varied in the entity classes studied, datasets have likewise differed in the scope of their tuple and entity annotations.
}
The most common format is \emph{$\langle$task, dataset, metric, score$\rangle$} (NLP-TDMS,~\cite{Hou2019Leaderboards}, PwC-LB~\cite{Kardas2020AXCELL}, TDMS-Ctx~\cite{kabongo2024effective}, SciLead~\cite{sahinuc2024efficient}), while the most comprehensive format is \emph{$\langle$task, dataset, metric, score, method$\rangle$} (LEGOBench, ~\cite{Singh2024LEGOBENCH}). 
These five datasets can be considered ``complete'' leaderboard datasets, as they include the score within the tuple.\footnote{These datasets can sometimes be divided into further subsets based on the size of the leaderboard. E.g., the ORKG-PwC and NLP-TDMS datasets filter out leaderboards with less than five entries.  Datasets can also be divided into pre-defined subsets. E.g., the ORKG datasets include pre-defined splits that correspond to experimentation by \citet{kabongo2024effective}\footnote{zero-shot versus few-shot, and the type of document representation: DocREC, DocTAET and DocFull}.
}
In contrast, two related datasets do not include scores: ORKG-PwC~\cite{Kabongo2021Mining}, and, SciREX~\cite{Jain2020SCIREX}.
Note that, for SciREX, the GitHub dataset includes a score.\footnote{\url{https://github.com/allenai/SciREX}} It is unclear whether this was added after the publication of the paper, demonstrating that data \textit{versioning} can be a challenge. 


\subsection{Source of Annotations}

Most datasets are assembled using manually curated leaderboards as a distant supervision source.  For example, the first leaderboard dataset, \emph{NLP-TDMS}~\cite{Hou2019Leaderboards}, was derived from a community-maintained GitHub repository \textit{NLP Progress}\footnote{\url{https://github.com/sebastianruder/NLP-progress}}, tracking state-of-the-art NLP datasets and tasks.
With the growing popularity of Paper with Code, many researchers turn to this resource to build ALG datasets, including ORKG-PwC, PwC-LB, SciREX, TDMS-Ctx and LEGOBench. 

{\color{orange}Not all datasets were created with manual annotations, however. Of the datasets derived from Papers with Code, only SciREX was subsequently corrected by a human annotator to ensure high accuracy.} Similarly, for SciLead~\cite{sahinuc2024efficient}, the leaderboard tuples $\langle$task, dataset, metric, score$\rangle$ were fully annotated by a single human annotator, {\color{orange}prioritising quality but limiting dataset size due to the manual effort involved.}

\subsection{Format of the Papers}

{\color{orange}Datasets differ in publication formats of the source publications. PDFs, though common, mix presentation with logical structure, whereas \LaTeX ~representations allows one to precisely isolate content from presentation. Some datasets use only one format, PDF (LEGOBench, SciLead) or \LaTeX(TDMS-Ctx), while others provide both (NLP-TDMS, ORKG-PwC, PwC-LB).} We note that this distinction is less important as tools like Grobid \cite{lopez2009grobid} grow in maturity to transform PDF files into a logical structure format, such as XML.

\section{{\color{blue}Overview of ALG Evaluation Metrics}}\label{sec:metrics}

One key issue in the field has been the use of various metrics for ALG evaluation, hindering result comparisons. Appendix~\ref{sec:metrics_details} lists all metrics used in leaderboard experiments. Below, we outline the key evaluation metrics used in prior work.

\subsection{Precision, Recall and F1} 

{\color{orange}Most work reports micro precision, recall, and F1, either for exact tuple matches or per entity class (e.g., task, metric). Some report macro variants, which offer deeper insights when frequent entities or tuples skew micro scores.
}

{\color{orange}Although not explicitly stated, we believe that generally
these scores are calculated per paper and then averaged.} However, \citet{Singh2024LEGOBENCH} calculated precision and recall per leaderboard. Experimental results can vary significantly depending on whether metrics are averaged across papers, leaderboards, or entities/tuples. {\color{blue}To demonstrate this significance, we replicated an experiment of \citet{sahinuc2024efficient} and found that if authors had used global averaging instead of per paper averages the recall would differ by 12.61.}\footnote{The authors conducted a zero-shot experiment evaluated using exact match. They reported a recall of 47.53 when averaging per paper, whereas the recall would have been 34.92 if averaged globally across all tuples.} In Table~\ref{tab:metrics} (Appendix~\ref{sec:metrics_details}), we provide definitions of these metrics.

{\color{orange}With the rise of generative AI with LLMs, there has been a need to explore string comparison metrics beyond exact match. For example, ~\citet{kabongo2024effective} explored partial matches. We note that metrics are useful in open-domain settings, where multiple valid expressions may exist and exact matching is too restrictive.}

\subsection{Leaderboard Specific Metrics}

{\color{orange}In addition to standard retrieval metrics, \citet{sahinuc2024efficient} introduced four metrics for leaderboard evaluation: leaderboard recall (LR), paper coverage (PC), result coverage (RC), and average overlap (AO). LR measures the percentage of correctly identified test leaderboards. PC and RC compute the average percentage of correctly linked papers and scores per leaderboard, respectively. AO quantifies the overlap between generated and test leaderboards~\cite{webber2010similarity}.}
{\color{blue}These leaderboard-specific metrics go beyond entity- or tuple-level evaluation by directly measuring the quality of the reconstructed leaderboard as a whole. This shift is crucial: standard precision and recall metrics may overlook whether the extracted information actually supports leaderboard reconstruction, i.e., better reflect the end-goal of ALG systems. Hence, adopting such metrics is essential for driving progress in building end-to-end usable and trustworthy leaderboard extraction tools.}





\subsection{Granularity of Science Concepts}

{\color{orange}As science advances, scientific concepts evolve. For example, broad terms like neural LMs may split into finer categories (e.g., \textit{pre-trained LMs} vs. \textit{LLMs}), or sibling concepts may merge or have their relative importance change (e.g., \textit{abstractive summarisation} overtaking \textit{extractive summarisation} as the dominant summarisation approach with the advent of deep learning). Similarly, what counts as an appropriate level of detail may change, such as the hyperparameters for neural networks, which has become part of standard reporting practice.
}





%
\subsection{{Extraction beyond Best Scores}}


{\color{orange}Current ALG’s focus on best scores limits its use to state-of-the-art comparisons and has drawn criticism for lacking real-world relevance. \citet{ethayarajh2020utility} highlight that this emphasis neglects factors like fairness, compactness, and energy efficiency. \citet{santy2021discussion} call for metrics beyond accuracy to better reflect practical utility. \citet{braggaar2024our} argue that rankings can mislead, as top models may underperform in practice. \citet{rodriguez2021evaluation} emphasise that not all evaluation examples are equally informative, urging leaderboards to account for difficulty. Together, these critiques advocate for broader, more meaningful evaluation.
}

{\color{blue}Including all experimental results introduces complexity, both methodologically (e.g., an LLM must extract more tuples, though many LLMs cannot output that many tokens) and from a user perspective (e.g., users must interpret a more complex leaderboard instead of a traditional one).}
\section{{\color{blue}ALG Unified Conceptual Framework}}
\label{sec:methods}

{\color{orange}To allow AI system builders to make system design choices based on research outcomes from ALG, we present the \emph{ALG Unified Conceptual Framework}.} For example, to build an ML leaderboard system, engineers may want to use the conceptualisation as inspiration for modules in a system architecture or agents in an Agentic AI system.

\begin{figure}
    \centering
    \includegraphics[width=.7\linewidth]{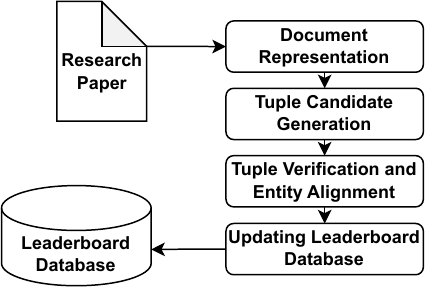}
    \caption{{\color{blue}ALG Unified Conceptual Framework.}}
    \label{fig:ideal_arch}
\end{figure}

{\color{orange}This conceptualisation is based on our analysis of the papers outlined in Table~\ref{tab:preliminaries}. Figure~\ref{fig:ideal_arch} illustrates these conceptual components and we provide examples of the methods for these components below, noting not all works include every component, reflecting differing research focuses.}

{\color{orange}The purpose of this conceptualisation is threefold: to (1) guide future researchers entering ALG research or building ALG systems; (2) organise the ALG experimentation space; and (3) understand the system-level importance of contributions.}

\subsection{Document Representation}
\label{sec:doc_rep}

{\color{orange}We note that several papers focus on finding the best representation of paper contents, whether starting from PDF or from structured formats like \LaTeX or XML. Such representations help highlight key information, especially when later ML components must process limited input text.}  

For example, approaches using pre-trained language models (e.g. BERT), document representation is crucial due to input length limits~\cite{Hou2019Leaderboards}. \citet{Hou2019Leaderboards} and \citet{Kabongo2021Mining} used \textit{document surrogates} like “DocTAET” (title, abstract, experimental setup, tables). Document representation can be more granular; for example, \citet{Jain2020SCIREX} use entity chains to detect tuples.

{\color{orange}Even with LLMs and their larger context windows, document representation remains important. Although LLMs can process full papers, the representation affects which information is used. \citet{kabongo2024effective}, for example, compare filtered document views with full-text inputs to assess effectiveness.}

An alternative document representation that should be explored in future research for ALG is the use of image-based inputs, where PDF pages are treated as images and processed by vision-based or multimodal LLMs. This preserves layout structure crucial for accurate table interpretation, especially in cases where leaderboard-relevant results are embedded in complex visual formats. Recent OCR-free models~\cite{khalighinejad2025matvix}, demonstrate that bypassing textual conversion of scientific text can reduce noise and improve tuple extraction accuracy in visually rich scientific documents.

To support large-scale ALG deployment, document representations like DocTAET can also be used to reduce LLM input length and inference cost by filtering to only the most relevant sections (e.g., title, abstract, tables). This makes processing more efficient when applied across tens of thousands of papers.

\subsection{Tuple Candidate Generation}\label{sec:cand_tuple}
{\color{orange}Given a document representation, this component extracts key contextual experimental attributes (e.g., task, dataset) and the result.} There are various ways to extract this information, based on how domain knowledge is used.

\subsubsection{Regarding Closed Domain Approaches}

For PTE closed domain approaches, entities are generally defined in a finite set (PTE class).  Any candidate tuples must be composed of these predefined entities and any new combination is acceptable.  For example, systems can  identify the key scientific concepts (e.g., extracting experiment attributes from relevant tables \cite{Kardas2020AXCELL}) to compose the tuples.  For PTT approaches, the aim is to match the predefined tuple with the source document, in order to check for an improvement in performance. \citet{Hou2019Leaderboards} frame this as a Natural Language Inference (NLI) task, to see whether the tuple is inferred by the document representation. 

\subsubsection{Regarding Open Domain Approaches}

{\color{orange}For open-domain approaches, tuples may include entities beyond a predefined list. For example, in SciREX~\cite{Jain2020SCIREX}, an entity detector identifies spans corresponding to task, data set, metric, or method. These unbounded entities are then used to compose tuples. However, the authors do not specify how the extracted tuples would update the leaderboard database.}

In \citet{sahinuc2024efficient}, detected entities correspond to concepts that fall into two categories: (1) unseen (i.e., new) and (2) seen.  Using a leaderboard database that is initially empty, entities are checked for corresponding entries, with either an exact match or a partial match.  If a match exists, the existing form in the database is used as the canonical representation for that concept.  This can be viewed as a data normalisation step. For all unmatched entities, these are treated as unseen, and a new database entry is created for them.

For the ALG data normalisation step, we recommend caching normalisation decisions: once the LLM maps entity or tuple A to B, the same rule can be reused for identical cases, avoiding repeated LLM calls and reducing computational cost.

\subsubsection{A Note on Score Extraction}
\label{sec:score_extraction}
Despite being central to ALG, only a handful of works \cite{Hou2019Leaderboards,Kardas2020AXCELL,Singh2024LEGOBENCH,kabongo2024effective,sahinuc2024efficient} extract best scores.  Other work focused on extracting the experimental conditions. We note that finding these conditions is a precursor to finding the full tuple for ALG (identifying experimental conditions to which the best score belongs). {\color{orange}For works that extract best scores, methods vary. \citet{Hou2019Leaderboards} apply heuristics based on orthographic features (boldface), whereas \citet{Kardas2020AXCELL} use more complex inferences, classifying table cells as numeric or non-numeric.}  Extracted quantities are normalised and the extreme (maximum or minimum) score is kept based on the metric.
{\color{orange}Earlier models used dedicated methods to align scores with conditions, whereas recent LLM prompting extracts entire tuples, including scores, with a single task-based prompt \cite{kabongo2024effective,Singh2024LEGOBENCH,sahinuc2024efficient}.}

\subsection{Tuple Verification and Entity Alignment}
\label{sec:verification_alignment}

{\color{orange}For each extracted tuple, the system should verify its correctness, especially for LLM-based approaches, which are susceptible to hallucination risk. Prior to the introduction of LLM, methods often implicitly included this step within the extraction process. For example, by framing the tuple generation task as an NLI problem, \citet{Hou2019Leaderboards} extract tuples that are aligned with the source content \textit{and} entailed by the source text, essentially providing some form of rationale for generated results. Others use partial alignment of the tuple at the entity level, such as using a Bayesian model to map different equivalent referring expressions to a canonical value \cite{Kardas2020AXCELL}.}

\subsection{Updating Leaderboard Database}
\label{sec:leaderboard_update}

{\color{orange}Once a tuple is verified, the final step is updating the leaderboard database. \citet{Kardas2020AXCELL} link experimental conditions to existing Papers with Code entries. Data may be normalised prior to this step~\cite{sahinuc2024efficient}, and filtered to exclude, for example, ablation studies~\cite{Kardas2020AXCELL}. Most prior work does \emph{not} detail this step, as the focus lies on NLP techniques for extraction rather than their downstream application, despite often being motivated by it.} Efficient ALG database updates at scale require batched writes and schema-aware indexing. To avoid redundant updates, tuples can be deduplicated with hash-based checks.

{\color{blue}

\section{ALG Benchmarking Guidelines}\label{sec:standardised_alg_benchmarking_guidelines}

\subsection{Open versus Closed Domain Reporting}

We recommend that researchers report results for both open- and closed-domain scenarios. Closed-domain, which assumes predefined entities and tuples, provides the simplest case and typically yields the highest accuracy. Open-domain, by contrast, does not rely on predefined knowledge and thus represents the most challenging case. However, in practical applications, scenarios will typically fall between these extremes. To ensure that benchmarking captures this full range of difficulty, and to allow comparisons across studies, we advise that researchers always include results for both domains. Including both allows to assess the feasibility of leaderboard extraction under both the most constrained and the most unconstrained settings, which reflects the diversity of real-world conditions.

\subsection{Dataset Reporting}

We recommend that researchers report results on publicly available datasets as a minimum requirement. We highlight \textbf{SciLead} and \textbf{LEGOBench} as two suitable options. SciLead is valuable for its fully human-curated annotations, ensuring high quality. LEGOBench offers the largest dataset with broad tuple coverage, enabling large-scale benchmarking across diverse tasks and methods. These two datasets are complementary: SciLead provides a gold standard for high-accuracy evaluation, while LEGOBench allows robust assessment at scale.  The feasibility of achieving broader and more informative evaluations strongly depends on ensuring open access to such datasets. Fortunately, SciLead and LEGOBench are fully open-source and thus support the practical feasibility of standardised evaluation without subscription or copyright barriers. However, a limitation of both datasets is that they only cover a restricted set of metadata attributes and focus solely on extracting the best results per paper. Therefore, in Section \S7.6, we recommend that researchers develop more comprehensive datasets that include all reported results and richer metadata. We also want to highlight the need for dataset versioning.  The documentation ambiguity for SciREX~\cite{Jain2020SCIREX}, as discussed in \S3.1, where the original paper omits the score while the GitHub repository later includes one, illustrates the problems which can arise when data differs from the original publication.

\subsection{Metrics}
Researchers should report precision, recall, and F1 as both \textbf{micro} and \textbf{macro} scores. Micro scores capture overall accuracy, favouring frequent entries, while macro scores weight papers, leaderboards, or entities equally and better reflect performance across varied result types. Reporting both provides balance, but most importantly researchers must clearly state the averaging method used (e.g. per paper, per leaderboard, or global).

In open-domain settings, exact string matching may be overly restrictive. We recommend reporting \textbf{partial match metrics}, which account for fuzzy or approximate matches. Such metrics better capture performance when multiple valid surface forms exist for the same scientific concept. This reflects real-world application scenarios more accurately.

To assess practical usability for leaderboard construction, researchers should report \textbf{leaderboard-specific metrics}. In particular, we highlight leaderboard recall (LR), paper coverage (PC), result coverage (RC), and average overlap (AO). These metrics provide insights into how effectively extracted tuples populate leaderboards. Leaderboard recall reflects whether leaderboards are correctly identified. Paper coverage measures whether all relevant papers are linked. Result coverage assesses the proportion of extracted results, and average overlap quantifies agreement between generated and ground truth leaderboards.

When possible, results should also be analysed across \textbf{fine-grained scientific concepts}. For example, extraction accuracy should be reported not only at the tuple level, but also separately for tasks, datasets, metrics, methods, and scores. This supports a nuanced understanding of performance, especially where new or rarely seen concepts may be difficult to extract.
}

\section{{\color{blue}ALG Challenges and New Directions}
}
\label{sec:reflection}

To help guide ALG researchers and system designers to potentially novel capabilities, {\color{blue}we list in this section challenges and new directions for ALG.}

\subsection{{New or Unseen Entities}}
{
The 2024 shared task on ALG \cite{d2024overview} highlights that many aspects of the task are still unsolved.  
{\color{orange}It includes closed and open domain subtasks, with the latter involving new entity detection.}\footnote{The organisers refer to these as \textit{few-shot} and \textit{zero-shot}, referring on current ML terminology.}  Indeed, \citet{Kabongo2023ZeroShot} showed that ML performance in extracting tuples with new entities (i.e., new scientific concepts, such as a newly introduced ML task or dataset) is much lower than extracting tuples with previously observed entities.} {\color{blue} In production, a challenge will be the feasibility of canonicalisation and disambiguation of these newly introduced ML entities. New entities often have ambiguous and inconsistent naming. For example, a newly introduced dataset might be referred to in short and long forms or with typos. In practice, feasibility depends on having automated canonicalisation methods that can cluster or align different surface forms of unseen entities. Without this, leaderboard entries will fragment into inconsistent records, undermining usability.}

\subsection{{Document Representation}}
{\color{orange}Representing source paper content remains an open challenge, even with LLMs' larger context windows. \citet{kabongo2024effective} found that using the full document with DocTAET led to worse tuple extraction, underscoring the need for representations that balance coverage and minimise irrelevant content during inference.} {\color{blue}Another practical feasibility consideration is that LLMs with larger context windows are more expensive, making it desirable for users to adopt document representations that allow feasible use of smaller, more efficient models.}

\subsection{{Extracting Numerical Scores}} 

{In most cases, the performance of tuple extraction, including scores, is significantly lower than that of tuples containing only the experimental conditions (which typically has F1 scores > 80), highlighting the difficulty of score extraction\cite{Kardas2020AXCELL,Hou2019Leaderboards,Yang2022TELIN,sahinuc2024efficient}.  For example, in recent work by \citet{sahinuc2024efficient}, score extraction using GPT-4  achieved an F1 score of approximately 70.}

{\color{blue}
Feasibility of extracting scores from a practical perspective goes further: not only must scores be extracted accurately, but extraction must be robust across various expressions of results. Systems must also handle ambiguous cases, such as ranges, averages, or multiple competing values. Current systems fall short in this respect, limiting the feasibility of fully automated leaderboard generation.}

{\color{blue}
\subsection{Feasibility of Extraction at Scale}

Most research papers benchmark ALG systems on dozens or hundreds of papers. However, production-grade leaderboards such as Papers with Code integrate tens of thousands of papers. Extracting tuples at this scale introduces feasibility challenges in computational efficiency and LLM inference cost. Practical implementation of an always-updating leaderboard requires optimised batching, caching strategies, and asynchronous processing.}

{\color{blue}
\subsection{Generalisability beyond ML}
A promising direction for future research is to explore the generalisability of ALG beyond ML. Domains like material science and biomedicine also report experimental results but use more varied formats and less standardised terminology. Key challenges include handling heterogeneous result expressions, complex domain language, and diverse contextual cues.
}

\subsection{{\color{blue}Comprehensive Leaderboards}}
{\color{blue}A key direction for future research is the development of comprehensive leaderboards. By comprehensive, we mean not only \emph{vertically}, by including all experimental results rather than only the best, but also \emph{horizontally}, by capturing richer metadata (e.g., hyperparameters). A necessary first step is the creation of a novel dataset to benchmark both existing and new techniques. } 
\section{Conclusion}\label{sec:conclusion}

{\color{blue}
In the position paper, we provide the \textbf{first overview of ALG research}, which reveals substantial diversity in problem framing and benchmarking practices. To address this fragmentation, we propose an \textbf{ALG unified conceptual framework} and present \textbf{ALG benchmarking guidelines}. Furthermore, our \textbf{first overview of ALG research to date} revealed that the scope of current leaderboards is limited. Therefore, one key recommendation in our list of \textbf{challenges and new directions for ALG} is to expand leaderboard coverage. Future leaderboards should report all results, including baselines, ablations, and method variations, and enrich tuples with broader metadata (e.g. hyperparameters) to create a more informative resource.} In support of this initiative, a continually updated reading list is maintained in a GitHub repository.\footnote{\url{https://github.com/RoelTim/ML-leaderboard-position-paper}}

\section*{Limitations}

A limitation of this paper is the scope, as we solely focus on the automatic generation of ML leaderboards. We note that other disciplines also report experimental outcomes, although the nature of the experimental procedures may differ. For example, \citet{ghosh2024toward} explore finetuning LLMs for schema-based information extraction in material science. Another example is \citet{wang2024scidasynth}, which introduce SciDaSynth, an interactive system using LLMs to extract and synthesise structured knowledge from the scientific literature in the form of tables.

While this position paper does not include new experiments, it aims to establish the foundational scaffolding required for rigorous future evaluations. We therefore provide clear evaluation setups and guidelines that, for future research, can be directly applied to assess existing and future ALG systems.

\section*{Ethics}

This research is subject to the governance by the ethics board of the Commonwealth Scientific and Industrial Research Organisation (CSIRO). We note that our proposal for AI research is to facilitate decision-making by users, as opposed to complete automation of tasks.  

\bibliographystyle{acl_natbib}
\bibliography{anthology,custom}


\appendix

\section{Acronyms}\label{app:acronyms}

\begin{table}[h]
\centering
\small
\begin{tabularx}{\linewidth}{lX}
\hline
\textbf{Acronym} & \textbf{Full form} \\
\hline
ALG & Automatic Leaderboard Generation \\
ML & Machine Learning \\
NLP & Natural Language Processing \\
LLM & Large Language Model \\
EMNLP & Conference on Empirical Methods in Natural Language Processing \\
PTE & Predefined Typed Entities \\
PTT & Predefined Typed Tuples \\
DocTAET & Document representation: Title, Abstract, Experimental Setup, Tables \\
DocREC & Document representation: Results, Experiments, Conclusion \\
ORKG & Open Research Knowledge Graph \\
PwC & Papers with Code \\
SciREX & Scientific Research Information Extraction (dataset) \\
LEGOBench & Leaderboard Generation Benchmark \\
SciLead & Scientific Leaderboard (dataset) \\
NLI & Natural Language Inference \\
F1 & Harmonic mean of Precision and Recall \\
LR & Leaderboard Recall \\
PC & Paper Coverage \\
RC & Result Coverage \\
AO & Average Overlap \\
\LaTeX & Document preparation system \\
OCR & Optical Character Recognition \\
\hline
\end{tabularx}
\caption{List of acronyms used in this paper.}
\label{tab:acronyms}
\end{table}

\section{Related Work Beyond ALG}
\label{sec:related_work}

\paragraph{Entity Recognition and Relation Extraction from Scientific Text}

Entity and relation extraction from scientific papers gained attention in 2017 with the SemEval-2017 ScienceIE task, which focused on identifying key elements like processes, tasks, and materials in publications~\cite{augenstein2017semeval}. The SemEval-2018 Task 7 advanced this by classifying relationships such as ``uses'', ``compares'', and 
``improves'' between scientific concepts~\cite{buscaldi2018semeval}.
\citet{mondal-etal-2021-end} built a knowledge graph from NLP papers by extracting four types
of relations: ``evaluatedOn'' (associating tasks with datasets), ``evaluatedBy'' (associating tasks with evaluation metrics), as well as  ``coreferent'' and ``related'' relations, which capture connections among entities of the same type.
Datasets like SciERC~\cite{luan2018multi}, TDMSci~\cite{Hou2021TDMSci}, and Dmdd~\cite{pan2023dmdd} further support entity extraction research.
The methods developed for scientific entity and relationship extraction can be leveraged to generate scientific leaderboards automatically.

\paragraph{Structured Scientific Information Extraction}

A scientific leaderboard compares methods, highlighting the best-performing one. It is a specific case of structured scientific information comparison and meta-analysis. Research has focused on extracting structured information without emphasising leaderboards. For example, \citet{pramanick-etal-2023-diachronic} leveraged a causal discovery algorithm to identify the TDMM (\emph{task, dataset, metric, method}) entities associated with a specific task  and assess
their causal influence on the task’s research trends.  Numerical quantity extraction has also been explored in scientific text for summarisation purposes \cite{10.1145/3539618.3591808}.

Work related to AI methods for publications from other scientific disciplines may also influence ALG.  \citet{ghosh2024toward} explored LLMs for schema-based information extraction in material science.
\citet{walker2023extracting} improved the extraction of experimental procedures using fine-tuned language models.  Other prior work has demonstrated the ability to extract methods and processes in scientific text for life sciences \cite{Wan2010SupportingSummariser, Wan2009WhettingContext} and material science \cite{yang-2022-piekm}.

\citet{wang2024scidasynth} introduced SciDaSynth, an interactive system using LLMs to extract and synthesise structured knowledge from scientific literature. Recently, \citet{pramanick-etal-2025-nature} proposed a method for automatically
extracting, categorizing, and quantitatively
analyzing contribution statements in research
papers. Knowledge graph generation has been explored in other disciplines, like astronomy \cite{10.1007/978-3-031-35320-8_15, timmer2023KG}.

\section{Problem Framing Details} 

Different methodologies for extracting leaderboard tuples rely on distinct document representations. The document representation defines which sections of a research paper are used before extracting leaderboard-related information. DocTAET contains text from a \textbf{Doc}ument’s \textbf{T}itle, \textbf{A}bstract, \textbf{E}xperimental Setup, and \textbf{T}able information. DocREC consists of text from a \textbf{Doc}ument’s \textbf{R}esults, \textbf{E}xperiments, and \textbf{C}onclusion sections. Some approaches extract content from the full paper, while others focus specifically on tables or citation tables. In Table~\ref{tab:docrep}, we show for each proposed methodology which document representation they use.

\begin{table}[ht!]
\setlength{\tabcolsep}{1pt}
\centering
\footnotesize
\begin{tabularx}{\linewidth}{ll}
\hline
\textbf{Methodology} & \makecell[l]{\textbf{Document}\\ \textbf{Representation}} \\
\hline
TDMS-IE~\cite{Hou2019Leaderboards}  & DocTAET\textsuperscript{*}, SC \\
ORKG-TDM \cite{Kabongo2021Mining} & DocTAET \\
ORKG-LB \cite{Kabongo2023ORKG}  &DocTAET \\
PI Graph \cite{singh2019automated} &Citation Tables \\
AxCell \cite{Kardas2020AXCELL} &Full Paper \& Tables\\
SciREX-IE \cite{Jain2020SCIREX} &Full Paper \\
TELIN \cite{Yang2022TELIN} & Full Paper \& Tables \\
TDMS-PR \cite{kabongo2024effective}  &  DocREC\textsuperscript{\textdagger} \\
MS-PR \cite{Singh2024LEGOBENCH}& Full Paper \\
TDMR-PR \cite{sahinuc2024efficient} &Full Paper \& Tables\\
\bottomrule
\addlinespace[1ex]

\multicolumn{2}{p{7cm}}{\textsuperscript{*} \citet{Hou2019Leaderboards} perform ablation studies with variations of DocTAET.}\\

\multicolumn{2}{p{7cm}}{\textsuperscript{\textdagger} \citet{kabongo2024effective} compare the performance of three document representations: DocREC, DocTAET, and the Full Paper.}\\
\end{tabularx}
\caption{\fontsize{10pt}{12pt}\selectfont Overview of the \textbf{Methodologies}. \textbf{Document Representation}: The content extracted from the paper before extracting the leaderboard tuples. }
\label{tab:docrep}
\end{table}

\section{Methodology Details}\label{sec:methodology_details}
\label{App:methodology_details}

In this section, we provide a summary of all the proposed ALG methodologies, and in Table~\ref{tab:nlp_models}, we list for each methodology which language models it uses.

\paragraph{TDMS-IE}

\citet{Hou2019Leaderboards} propose TDMS-IE, a methodology to automatically extract \(\langle\)task, dataset, metric, score\(\rangle\) tuples from research papers. The first step of TDMS-IE is extracting the document representation and the score context from the research paper. The document representation, DocTAET, covers the title, abstract, experimental setup, and table information. The title and abstract help predict the task, while the experimental setup and table information assist in identifying the dataset and metric.  A second document-based structure, the score context, SC, represents contents from tables, since the work relies on table-based (and formatting, i.e., bold font) heuristics to generate candidate tuples.  The SC captures the table caption and column headers corresponding to each bold-faced numeric score in each table of the research paper. This is used in conjunction with formatting-based heuristics to identify candidates for the best score of a \(\langle\)task, dataset, metric\(\rangle\) tuple.\footnote{For example, bold-faced scores are most likely to be best score.} \citet{Hou2019Leaderboards} frame the problem as a natural language inference (NLI) task using two entailment models: 1) DocTAET-TDM and 2) SC-DM.  Each model generates a tuple hypothesis (a Task-Dataset-Metric, or TDM, tuple for DocTAET-TDM; a Score-Dataset-Metric tuple for SC-DM), by searching for candidate argument combinations from a ``taxonomy'' (that is, a knowledge base) of previously observed tuples. A fine-tuned BERT model (for NLI) predicts whether a candidate tuple can be inferred from DocTAET, inferring links between the paper’s text and the predefined canonical labels for the \textit{Task, Dataset, and Metric}, as represented in the taxonomy. For instance, the model can recognise that "Rg-2" and "ROUGE-2" refer to the same metric. Similarly, 
the SC-DM infers entailment relationships between the SC document representations and dataset-metric tuples. Both models use the BERT model limited to 512 tokens~\cite{devlin2019bert}, although newer models with larger token capacities may improve performance.

\begin{table*}[tbh]
\setlength{\tabcolsep}{1pt}
\centering
\small
\begin{tabularx}{\linewidth}{lX}
\hline
\textbf{Methodology} & \textbf{Language Models} \\
\hline
TDMS-IE  \cite{Hou2019Leaderboards} & BERT \cite{devlin2019bert} \\
ORKG-TDM  \cite{Kabongo2021Mining} & XLNet~\cite{yang2019xlnet}, SciBERT~\cite{beltagy2019scibert}, BERTbase~\cite{devlin2019bert} \\
ORKG-LB  \cite{Kabongo2023ORKG} & BERT~\cite{devlin2019bert}, SciBERT~\cite{beltagy2019scibert}, XLNet~\cite{yang2019xlnet}, BigBird~\cite{michalopoulos-etal-2022-icdbigbird} \\
PI Graph  \cite{singh2019automated} & Undefined \\
AxCell  \cite{Kardas2020AXCELL} & ULMFiT classifier~\cite{howard2018ulmfit}, BM25~\cite{robertson2009probabilistic} \\
SciREX-IE  \cite{Jain2020SCIREX} & SciBERT~\cite{beltagy2019scibert}, BiLSTM~\cite{graves2005framewise} \\
TELIN  \cite{Yang2022TELIN} & SpERT~\cite{eberts2020span} \\
TDMS-PR  \cite{kabongo2024effective} & Llama 2~\cite{touvron2023llama}, Mistral~\cite{jiang2023mistral} \\
MS-PR  \cite{Singh2024LEGOBENCH} & Falcon~\cite{almazrouei2023falcon}, Galactica~\cite{taylor2022galactica}, Llama 2~\cite{touvron2023llama}, Llama 3~\cite{dubey2024llama}, Mistral~\cite{jiang2023mistral}, Vicuna~\cite{chiang2023vicuna}, Zephyr~\cite{tunstall2023zephyr}, Gemini~\cite{team2023gemini}, GPT-4~\cite{achiam2023gpt}  \\
TDMR-PR  \cite{sahinuc2024efficient} & Llama 2~\cite{touvron2023llama}, Llama 3~\cite{dubey2024llama}, Mixtral~\cite{jiang2024mixtral}, GPT-4~\cite{achiam2023gpt} \\
\bottomrule
\addlinespace[1ex]
\end{tabularx}
\caption{\fontsize{10pt}{12pt}\selectfont Overview of the language models used in each methodology, demonstrating how the methodologies have (logically) adopted more advanced models over time as discussed in Section~\ref{sec:methods}.}
\label{tab:nlp_models}
\end{table*}

\paragraph{PI Graph}
\citet{singh2019automated} introduce the performance improvement graph (PI Graph) to rank research papers based on their performance. This graph is constructed from \emph{performance tables}, which compare the methodologies and results of a paper with those from previous works. Citations within these tables create edges between papers, reflecting performance improvements. However, the authors do not detail how the performance tables are identified, extracted, or processed. The focus of this work is on ranking papers by performance, not on the extraction of leaderboard tuples, which falls outside the scope of their methodology.

\paragraph{AxCell}
\citet{Kardas2020AXCELL} introduce AxCell, a pipeline for automatically extracting results from machine learning papers. AxCell first categorises tables into leaderboard, ablation, or irrelevant types using the ULMFiT classifier \cite{howard2018ulmfit}. For leaderboard and ablation tables, each cell is classified as a dataset, metric, paper model, cited model, or other. BM25 \cite{robertson2009probabilistic} is employed to extract relevant context from the paper for each cell. A generative model, based on the naive Bayes assumption, then links numeric cells to predefined leaderboards. Finally, the system filters out cited models, low-scoring links, and inferior results, retaining only the top results for each leaderboard.

\paragraph{SciREX-IE}
\citet{Jain2020SCIREX} introduce SciREX-IE, a methodology for extracting N-ary relations from research papers. The process starts by extracting raw text and section information from documents (excluding figures, tables, and equations). SciREX-IE encodes the text in two steps: first, section-level token embeddings are obtained using SciBERT \cite{beltagy2019scibert}, followed by a BiLSTM \cite{graves2005framewise} to capture cross-section dependencies. A BIOUL-based CRF tagger identifies and classifies mentions using BERT-BiLSTM embeddings, which are created by combining token embeddings with additional features. The system classifies mentions as salient or not and performs coreference resolution using the SciBERT embeddings, clustering mentions into entities. Salient clusters are then used for relation extraction, with document-level embeddings aggregating section data. The model jointly optimises mention identification, saliency classification, and relation extraction during training.

\paragraph{ORKG-TDM}
\citet{Kabongo2021Mining} propose ORKG-TDM, a methodology to extract \(\langle\)task, dataset, metric\(\rangle\) tuples from research papers. The authors refer to their approach as the ORKG-TDM, as it is integrated into a scholarly knowledge platform called Open Research Knowledge Graph (ORKG)~\cite{jaradeh2019open}. ORKG-TDM follows a similar approach to TDMS-IE \cite{Hou2019Leaderboards} by framing the tuple extraction problem as an entailment problem, but uses a single-step approach. As in TDMS-IE, DocTAET is the document representation, and leaderboard tuples coming from a predefined taxonomy are the hypotheses. New to ORKG-TDM is a task-specific parameter for the number of false triples per paper. While \citet{Hou2019Leaderboards} conducted experiments with only the original BERT model for TDMS-IE, \citealt{Kabongo2021Mining}, in implementing the ORKG-TDM methodology, also experimented with the pre-trained SciBERT model~\cite{beltagy2019scibert}, designed for scientific text, and XLNet~\cite{yang2019xlnet}, an autoregressive transformer capable of handling contexts longer than BERT's 512-token maximum.

\paragraph{TELIN}
\citet{Yang2022TELIN} proposed TELIN, a methodology to extract \(\langle\)task, dataset, model, method\(\rangle\) tuples from research papers. TELIN begins by converting unstructured PDFs into structured documents, using YOLO to detect paragraphs, headings, captions, and tables \citep{redmon2016you}. SPLERGE is then applied to extract table components such as rows, columns, and cells \citep{tensmeyer2019deep}. For NER, TELIN uses SpERT, a BERT-based model pre-trained on the SCiERC dataset, to classify scientific entities into categories like task, method, dataset, and evaluation metric \citep{eberts2020span}. String matching between these entities and non-numeric table cells is performed using fuzzy search to handle non-exact matches and acronyms. Tuples are formed when at least three of the four entities (task, dataset, metric, model) are identified within the table and its caption. These extracted leaderboards are stored in a shared knowledge base, which is iteratively refined to discover more entities across documents. A human review stage prioritises uncertain entities, using feedback to fine-tune SpERT, iterating until entity prediction stabilises.

\paragraph{ORKG-LB}
\citet{Kabongo2023ORKG} introduced ORKG Leaderboard (ORKG-LB), a follow-up methodology of ORKG-TDM \cite{Kabongo2021Mining}. ORKG-LB focuses on the extraction of the \(\langle\)task, dataset, metric\(\rangle\) tuples by framing the extraction task as an entailment problem. ORKG-LB starts by allowing users to input a LaTeX or PDF version of the research paper. ORKG-LB uses the GROBID parser~\cite{lopez2009grobid} for PDF files and PANDOC~\cite{pandoc} to convert LaTeX files into XML TEI markup. Then, ORKG-LB extracts DocTAET~\cite{Hou2019Leaderboards}, focusing on sections likely to contain task–dataset–metric mentions, reducing noise and enhancing generalisation. For training the inference, for each paper, positive and negative samples of tuples are required. For the number of false triples per paper, ORKG-LB relies on the same task-specific parameter as used for ORKG-TDM. For the inference model, the authors of ORKG-LB experiment with four different transformer model variants: BERT~\cite{devlin2019bert}, SciBERT~\cite{beltagy2019scibert}, XLNet~\cite{yang2019xlnet} and BigBird~\cite{zaheer2020big}.

\paragraph{TDMS-PR}
The work of \citet{kabongo2024effective} experiments with prompting LLMs to extract \(\langle\)task, dataset, metric, score\(\rangle\) tuples from research papers, and we refer to this methodology as TDMS-PR. The authors experiment with different document representations provided to the LLM when prompting the LLM. They propose a novel document representation, DocREC, which comprises text from the results (R), experiments (E) and conclusions (C) sections. They compare the results when using DocREC to when using DocTAET \cite{Hou2019Leaderboards} or DocFull, which is the full paper as document representation. On average, DocREC consists of more tokens than DocTAET, 1,586 versus 493, and by definition, DocFull is by default always the longest document representation. The authors experiment with LLMs from the Flan-T5 collection, Mistral 7B and Llama 3 7B.

\paragraph{MS-PR}
The authors of \citet{Singh2024LEGOBENCH} prompt an LLM to extract the \(\langle\)method, score\(\rangle\) tuple given a research paper representation and a \(\langle\)task, dataset, metric\(\rangle\) tuple; we refer to this as MS-PR. While both TDMS-PR~\cite{kabongo2024effective} and MS-PR are prompt-based, their tuple scopes differ: TDMS-PR focuses on \(\langle\)task, dataset, metric\(\rangle\), while MS-PR targets \(\langle\)method, score\(\rangle\). \citet{Singh2024LEGOBENCH} experiment with MS-PR by using a wide range of LLMs: Falcon, Falcon Instruct, Galactica, Llama 2 (7B \& 13B), Llama 2 Chat (7B \& 13B), Mistral Instruct, Vicuna (7B \& 13B), Zephyr Beta, Gemini Pro and GPT-4~\cite{almazrouei2023falcon, taylor2022galactica, touvron2023llama, jiang2023mistral, chiang2023vicuna, zephyr_beta, team2023gemini, achiam2023gpt}.

\paragraph{TDMR-PR}
The authors of \citet{sahinuc2024efficient} prompt an LLM to extract \(\langle\)task, dataset, metric, score\(\rangle\) tuples, we refer to this method as TDMR-PR. First, TDMR-PR extracts the tuples from the papers via a retrieval-augmented generation method using an LLM. Second, depending on the domain (closed, hybrid, or, open), TDMR-PR normalises these tuples to a predefined taxonomy or creates new entries for novel tasks, datasets, or metrics. Lastly, TDMR-PR ranks the papers based on their performance, constructing or updating leaderboards accordingly.

\section{Dataset Details}\label{sec:dataset_details}

Table~\ref{tab:datasets_extended} presents an extended version of Table~\ref{tab:datasets}, providing detailed information for each version of the included datasets. For every train, test, and validation split, we report the number of associated papers and extracted tuples. This table highlights the substantial diversity across datasets, which complicates direct comparisons between experiments.
    
\setlength{\tabcolsep}{1.5pt}
\begin{table*}[ht]
\centering
\fontsize{9pt}{8pt}\selectfont
\begin{tabularx}{\linewidth}{llccccccccccccccccc}

 \toprule
&\textbf{}& \multicolumn{5}{c}{\textbf{Entities}} & \multicolumn{2}{c}{\textbf{Format}} & \multicolumn{3}{c}{\textbf{Annotations\textsuperscript{*}}}&\textbf{Unk.}&\multicolumn{2}{c}{\textbf{Train Stats.}}&\multicolumn{2}{c}{\textbf{Test Stats.}}&\multicolumn{2}{c}{\textbf{Val. Stats.}}\\

\cmidrule(lr){3-7} \cmidrule(lr){8-9} \cmidrule(lr){10-12} \cmidrule(lr){14-15} \cmidrule(lr){16-17} \cmidrule(lr){18-19}

\textbf{Paper} & \makecell[l]{\textbf{V}} & \textbf{T} & \textbf{D} & \textbf{M} & \textbf{S} & \textbf{Md} & \textbf{PDF} & \textbf{\LaTeX} & \textbf{HA} & \textbf{PwC} & \textbf{NLPP} &\makecell[l]{\textbf{Ann.}}&\textbf{\#P} &\textbf{\#T} &\textbf{\#P} & \textbf{\#T}&\textbf{\#P} & \textbf{\#T} \\

 \midrule  
\multicolumn{19}{l}{\textbf{ORKG-PwC Dataset}}\\
\citet{Kabongo2021Mining} & {v1} & \cmark & \cmark & \cmark & \xmark & \xmark & \cmark & \xmark & \xmark & \cmark & \xmark &\xmark &2,831\textdagger &11,724\textdagger&1,228\textdagger& 5,060\textdagger &-&- \\
\citet{Kabongo2021Mining} & {v2} & \cmark & \cmark & \cmark & \xmark & \xmark & \cmark & \xmark & \xmark & \cmark & \xmark &\cmark &3,753\textdagger &11,724\textdagger& 1,608\textdagger& 5,060\textdagger &-&-\\

\citet{Kabongo2023ORKG}& {v3} & \cmark & \cmark & \cmark& \xmark & \xmark&\xmark&\cmark& \xmark&\cmark&\xmark&\xmark&587\textdagger &9,614\textdagger &270\textdagger &4,096\textdagger &-&-\\
\citet{Kabongo2023ORKG}& {v4} & \cmark & \cmark & \cmark& \xmark & \xmark&\xmark&\cmark& \xmark&\cmark&\xmark&\cmark&2,946\textdagger& 9,614\textdagger &1,262\textdagger &4,096\textdagger &-&-\\

\citet{Kabongo2023ORKG}& {v5} & \cmark & \cmark & \cmark& \xmark & \xmark&\cmark&\xmark&\xmark&\cmark&\xmark&\xmark& 587\textdagger &9,614\textdagger &270\textdagger &4,096\textdagger &-&-\\
\citet{Kabongo2023ORKG}& {v6} & \cmark & \cmark & \cmark& \xmark & \xmark&\cmark&\xmark&\xmark&\cmark&\xmark&\cmark&2,946\textdagger &9,614\textdagger &1,262\textdagger &4,096\textdagger &-&-\\

\citet{Kabongo2023ZeroShot}& v7\textsuperscript{\#} & \cmark & \cmark & \cmark& \xmark & \xmark&\cmark&\xmark& \xmark&\cmark&\xmark&\cmark&-&-&1,000&1,925&-&-\\
 \midrule  
\multicolumn{19}{l}{\textbf{NLP-TDMS Dataset}}\\
\citet{Hou2019Leaderboards} &{v1} & \cmark & \cmark & \cmark & \cmark & \xmark & \cmark & \xmark & \xmark & \xmark & \cmark & \xmark &124&325&118&281 &-&-\\
\citet{Hou2019Leaderboards} & {v2} & \cmark & \cmark & \cmark & \cmark & \xmark & \cmark & \xmark & \xmark & \xmark & \cmark &\cmark & 170&325&162&281&-&- \\
\citet{Kardas2020AXCELL} & {v3} & \cmark & \cmark & \cmark & \cmark & \xmark & \xmark & \cmark & \xmark& \xmark&\cmark&\cmark& \(\leq\)170&\(\leq\)325&\(\leq\)162&\(\leq\)281&-&- \\
 \midrule  
\multicolumn{19}{l}{\textbf{PwC-LB Dataset}}\\
\citet{Kardas2020AXCELL}&{v1} & \cmark & \cmark & \cmark & \cmark & \xmark & \xmark & \cmark &\xmark&\cmark&\xmark &\xmark & \textdaggerdbl & \textdaggerdbl & 516 & 2,802& \textdaggerdbl &\textdaggerdbl \\

\citet{Yang2022TELIN} & {v2} & \cmark & \cmark & \cmark & \cmark & \xmark& \cmark&\xmark&\xmark&\cmark&\xmark &\xmark&- & -& 516 & 2,802 &-&-\\
 \midrule  
\multicolumn{19}{l}{\textbf{SciREX Dataset}}\\
\citet{Jain2020SCIREX} & & \cmark & \cmark & \cmark & \xmark & \cmark & \(\sim\) & \(\sim\) & \cmark & \cmark & \xmark & \xmark&\(\leq\)438 & \(\nabla\)&\(\leq\)438 & \(\nabla\)& \(\leq\)438 & \(\nabla\)\\
 \midrule  
\multicolumn{19}{l}{\textbf{TDMS-Ctx Dataset}}\\
\citet{kabongo2024effective}&v1\textsuperscript{\textsection}&\cmark&\cmark&\cmark&\cmark&\xmark&\xmark&\cmark&\xmark&\cmark&\xmark&\cmark&{11,807}&402,409&{1,326}&33,863&-&-\\ 
\citet{kabongo2024effective}&v2\textsuperscript{\textsection}&\cmark&\cmark&\cmark&\cmark&\xmark&\xmark&\cmark&\xmark&\cmark&\xmark&\cmark&{12,388}&415,788&{1,401}&34,799&-&-\\ 
\citet{kabongo2024effective}&v3\textsuperscript{\textsection}&\cmark&\cmark&\cmark&\cmark&\xmark&\xmark&\cmark&\xmark&\cmark&\xmark&\cmark&{10,058}&415,788&{1,105}&31,213&-&-\\ 
\citet{kabongo2024effective}&v4\textsuperscript{\textsection}&\cmark&\cmark&\cmark&\cmark&\xmark&\xmark&\cmark&\xmark&\cmark&\xmark&\cmark&{11,807}&402,409&{746}&14,604&-&-\\ 
\citet{kabongo2024effective}&v5\textsuperscript{\textsection}&\cmark&\cmark&\cmark&\cmark&\xmark&\xmark&\cmark&\xmark&\cmark&\xmark&\cmark&{12,388}&415,788&{789}&14,800&-&-\\ 
\citet{kabongo2024effective}&v6\textsuperscript{\textsection}&\cmark&\cmark&\cmark&\cmark&\xmark&\xmark&\cmark&\xmark&\cmark&\xmark&\cmark&{10,058}&415,788&{595}&14,273&-&-\\ 
 \midrule  
\multicolumn{19}{l}{\textbf{LEGOBench Dataset}}\\
\citet{Singh2024LEGOBENCH}&&\cmark&\cmark&\cmark&\cmark&\cmark&\cmark&\xmark&\xmark&\cmark&\xmark&\cmark&-&-&\(\lozenge\) & 43,105&-&-\\
 \midrule  
\multicolumn{19}{l}{\textbf{SciLead Dataset}}\\
\citet{sahinuc2024efficient} & & \cmark & \cmark & \cmark & \cmark & \xmark &\cmark & \xmark & \cmark & \xmark & \xmark & \xmark&- &-& 43 &\(\oslash\)&-&-\\
\bottomrule
\addlinespace[1ex]
\multicolumn{19}{p{16cm}}{
* For annotations, we distinguish between human annotations (HA), Papers with Code (PwC) and NLP Progress (NLPP), however PwC includes partial human annotation, and domain experts fully curated NLP Progress via GitHub pull requests. 
\(\sim\) Use LaTeX if available; otherwise, default to PDF. 
\textdaggerdbl Different data for training (unlabelled arXiv papers and segmented tables) and validation (linked results). 
\textdagger Two-fold cross-validation: 70\% train, 30\% test, with averaged results.
\textsuperscript{\(\lozenge\)} 9,847 leaderboards, and the number of papers is unspecified.
\textsuperscript{\textsection} v1–v3 are few-shot experiment datasets with document representations: v1 (DocFULL), v2 (DocREC), and v3 (DocTAET). v4–v6 are zero-shot experiment datasets with the same representations: v4 (DocFULL), v5 (DocREC), and v6 (DocTAET). \textsuperscript{\#} the same data source as v2, but with updated timestamps and no overlap with v2. \(\nabla\) An average of 5 tuples annotations per paper. \(\oslash\) Unspecified, with 138 unique tuples reported.
}\\
\end{tabularx}
\vspace{-2pt}
\caption{\fontsize{10pt}{12pt}\selectfont This table summarises the datasets from multiple research papers, detailing dataset variant (\textbf{V}), \textbf{Entities} captured (\textbf{T} = Task, \textbf{D} = Dataset, \textbf{M} = Metric, \textbf{S} = Score, \textbf{Md} = Method), \textbf{format} (PDF, \LaTeX), \textbf{Annotations} (\textbf{HA} = Human Annotation, \textbf{PwC} = Papers with Code, \textbf{NLPP} = NLP Progress), and inclusion of unknown annotations (\textbf{Unk. Ann.}). Additionally, the table includes \textbf{Train}, \textbf{Test}, and validation (\textbf{Val.}) statistics (\textbf{Stats.}): the number of papers (\textbf{\#P}) and tuples (\textbf{\#T}).}
\label{tab:datasets_extended}
\end{table*}

\section{Definitions of Metrics}\label{sec:metrics_details}

\begin{table*}[tbh]
\setlength{\tabcolsep}{2pt} 
\centering
\footnotesize
\begin{tabularx}{.75\linewidth}{lccccccccX}
\hline
 & \multicolumn{3}{c}{\textbf{Micro}} & \multicolumn{3}{c}{\textbf{Macro}} & \multicolumn{2}{c}{\textbf{Part. Micro}} & \\
\cmidrule(lr){2-4} \cmidrule(lr){5-7} \cmidrule(lr){8-9}
\textbf{Paper} & \textbf{P} & \textbf{R} & \textbf{F1} & \textbf{P} & \textbf{R} & \textbf{F1} & \textbf{P} & \textbf{F1} & \textbf{Other Metrics} \\
\hline
\citet{Hou2019Leaderboards} & \cmark  & \cmark &\cmark  &\cmark  & \cmark  & \cmark  & \xmark  & \xmark& None \\
\citet{Kabongo2021Mining} &\cmark  & \cmark &\cmark  &\cmark  & \cmark  & \cmark  & \xmark  & \xmark& None \\
\citet{Kabongo2023ORKG}&\cmark  & \cmark &\cmark  &\cmark  & \cmark  & \cmark  & \xmark  & \xmark& None \\
\citet{Kabongo2023ZeroShot} &\cmark  & \cmark &\cmark  &\cmark  & \cmark  & \cmark  & \xmark  & \xmark& None \\
\citet{Kardas2020AXCELL} &\cmark  & \cmark &\cmark  &\cmark  & \cmark  & \cmark  & \xmark  & \xmark& None \\
\citet{Jain2020SCIREX} &  \cmark  & \cmark &\cmark  & \xmark & \xmark & \xmark & \xmark & \xmark & None \\
\citet{Yang2022TELIN} &\cmark  & \cmark &\cmark  &\cmark  & \cmark  & \cmark  & \xmark  & \xmark& None \\
\citet{kabongo2024effective} &  \cmark & \xmark & \cmark &  \xmark & \xmark & \xmark & \cmark & \cmark & None \\
\citet{Singh2024LEGOBENCH} & \cmark &\cmark & \xmark & \xmark & \xmark & \xmark & \xmark & \xmark & None\\
\citet{sahinuc2024efficient} &\cmark & \cmark & \cmark & \xmark & \xmark & \xmark & \xmark & \xmark &leaderboard recall (LR), paper coverage (PC), result coverage (RC), and average overlap (AO) \\
\bottomrule
\end{tabularx}
\caption{\fontsize{10pt}{12pt}\selectfont Overview of evaluation metrics used in each paper.}
\label{tab:metrics}
\end{table*}

In this section, we define the micro and macro versions of the Precision, Recall, and F1 metrics for the ALG task. Based on our best guess, most of the existing works typically compute micro precision, micro recall, and micro F1 by first calculating these scores per paper and then averaging them. However, this is solely a best guess, and we know that, for example, \citet{kabongo2024effective} and \citet{Singh2024LEGOBENCH} calculate the score on a leaderboard level. We recommend that future researchers either use these definitions of these metrics or explicitly specify if they average across a different dimension (e.g., across leaderboards), as the choice of the averaging method can significantly impact the final score.

\begingroup
\small
\vspace{-5pt} \begin{align} \text{Micro P} = \frac{1}{P} \sum_{p=1}^{P} \frac{\sum_{i=1}^{N_p} TP_{p,i}}{\sum_{i=1}^{N_p} (TP_{p,i} + FP_{p,i})} \end{align} \vspace{-5pt}
\endgroup

where \( P \) represents the total number of papers, and \( N_p \) represents the total number of extracted leaderboard tuples or entities, per paper \( p \). The term \( TP_{p,i} \) denotes the number of true positive instances for the \( i \)-th instance in paper \( p \), while \( FP_{p,i} \) represents the number of false positive instances for the \( i \)-th instance in the same paper. The precision is first computed for each individual paper before being averaged across all \( P \) papers. 


Micro Recall measures the proportion of correctly identified leaderboard entities/tuples:

\begingroup
\small
\vspace{-5pt} \begin{align} 
\text{Micro R} &= \frac{1}{P} \sum_{p=1}^{P} \frac{\sum_{i=1}^{N_p} TP_{p,i}}{\sum_{i=1}^{N_p} (TP_{p,i} + FN_{p,i})} 
\end{align} \vspace{-5pt}
\endgroup

where \( FN_{p,i} \) represents the number of false negatives for the \( i \)-th instance in paper \( p \). 

Micro F1 is the \emph{harmonic} mean of micro precision and micro recall, providing a balanced measure of extraction performance:

\begingroup
\small
\vspace{-5pt} \begin{align} 
\text{Micro F1} &= \frac{2 \times \text{Micro P} \times \text{Micro R}}{\text{Micro P} + \text{Micro R}} 
\end{align} \vspace{-5pt}
\endgroup

We recommend also reporting the macro variants of these metrics to give more insight if some of the entries/tuples appear frequently and, therefore, disproportionally influence the micro scores. For \emph{macro} metrics, we first average across all classes and then across \( P \) papers. Macro precision is given by:

\begingroup
\small
\vspace{-5pt} \begin{align} 
\text{Macro P} = \frac{1}{P} \sum_{p=1}^{P} \frac{1}{C_p} \sum_{c=1}^{C_p} \frac{\sum_{i=1}^{N_{p,c}} TP_{p,c,i}}{\sum_{i=1}^{N_{p,c}} (TP_{p,c,i} + FP_{p,c,i})} 
\end{align} \vspace{-5pt}
\endgroup

where \( C_p \) is the number of classes for each paper \(p\).

Macro Recall is given by:

\begingroup
\small
\vspace{-5pt} \begin{align} 
\text{Macro R} = \frac{1}{P} \sum_{p=1}^{P} \frac{1}{C} \sum_{c=1}^{C} \frac{\sum_{i=1}^{N_{p,c}} TP_{p,c,i}}{\sum_{i=1}^{N_{p,c}} (TP_{p,c,i} + FN_{p,c,i})} 
\end{align} \vspace{-5pt}
\endgroup

And Macro F1 is given by:

\begingroup
\small
\vspace{-5pt} \begin{align} 
\text{Macro F1} = \frac{1}{P} \sum_{p=1}^{P} \frac{1}{C} \sum_{c=1}^{C} \frac{2 \times \text{P}_{p,c} \times \text{R}_{p,c}}{\text{P}_{p,c} + \text{R}_{p,c}} 
\end{align} \vspace{-5pt}
\endgroup

It is important to note that these definitions serve as an example of how micro and macro variations can be calculated when averaged at the paper level. However, these definitions can be easily adapted for calculations at the leaderboard level.

\section{An Overview of Experimental Results}\label{sec:app_exp_results}

We have compiled all the results we could find in the literature where researchers experiment with extracting leaderboard tuples and entities, evaluating these extractions using micro, partial micro, or macro precision, recall, and F1 scores. Tables~\ref{tab:task-dataset-metric} -~\ref{tab:method} present an overview of these experiments. \textbf{These tables highlight the complexity of comparing different results due to the diversity of problem framing (e.g. closed versus open domain), datasets and metrics.} We omitted details on how the scores were averaged (e.g., across papers or leaderboards), as this information is often not reported in many studies. These differences in averaging methods also complicate direct comparisons between works. Please note that there may be additional subtle variations in the experimental setup that are not captured in these tables, which could prevent a fair comparison.

\onecolumn

\begin{table*}

\setlength{\tabcolsep}{1.0pt}
\fontsize{8pt}{10pt}\selectfont 

\renewcommand{\arraystretch}{0.9}
\begin{tabular}{l N N N N N N N N L{2.2cm} L{2cm} p{2.8cm}}
\toprule
&\multicolumn{3}{c}{\textbf{Micro}} & \multicolumn{3}{c}{\textbf{Macro}}& \multicolumn{2}{c}{\textbf{Part. Micro}} &&\\
\cmidrule(lr){2-4} \cmidrule(lr){5-7} \cmidrule(lr){8-9} 
\textbf{Reported In}& \textbf{P} &\textbf{R} & \textbf{F1} & \textbf{P} & \textbf{R} & \textbf{F1} & \textbf{P} & \textbf{F1} & \textbf{Dataset} & \textbf{Method}& \textbf{Experimental Setup}\\
\bottomrule
 \multicolumn{12}{l}{\rule{0pt}{1.5ex} \textbf{Closed Domain Problem Framing}}\\[0.1ex]
\toprule

\citet{Hou2019Leaderboards} & 60.2& 73.1 & 66.0 & 54.1 & 65.9 & 56.6 &&& {NLP-TDMS-v1} & TDMS-IE&\\
\citet{Hou2019Leaderboards}&29.4&42.0&34.6&24.9&43.6&28.1 &&&{NLP-TDMS-v1} &EL{\textsuperscript{\textdagger}}&\\
\citet{Hou2019Leaderboards}& 56.8&23.8&33.6&56.8&30.9&37.3 &&&{NLP-TDMS-v1} &MLC{\textsuperscript{\textdagger}}&\\
\citet{Hou2019Leaderboards}&16.8&7.8&10.6& 8.1&6.4&6.9&&&{NLP-TDMS-v1} &SM{\textsuperscript{\textdagger}}&\\

\citet{Hou2019Leaderboards} & 60.8 & 76.8 & 67.8 & 62.5 & 75.2 & 65.3 &&&{NLP-TDMS-v2} & TDMS-IE&\\
\citet{Hou2019Leaderboards}&24.3&36.3&29.1&18.1&31.8&20.5 &&&{NLP-TDMS-v2} &EL{\textsuperscript{\textdagger}}&\\
\citet{Hou2019Leaderboards}&42.0&20.9& 27.9&42.0&23.1&27.8&&&{NLP-TDMS-v2} &MLC{\textsuperscript{\textdagger}}&\\
\citet{Hou2019Leaderboards}& 36.0&19.6&25.4& 31.8&30.6&31.0&&&{NLP-TDMS-v2} &SM{\textsuperscript{\textdagger}}&\\
\citet{Hou2019Leaderboards} & 68.6&40.3 &50.8 & 29.6& 29.1 & 28.1 &&&{NLP-TDMS-v2}& TDMS-IE&TAE\textsuperscript{\#}\\
\citet{Hou2019Leaderboards} & 50.0&23.7 &32.2 & 20.8& 20.1&19.4 &&&{NLP-TDMS-v2} & TDMS-IE&TAT\textsuperscript{\#}\\
\citet{Hou2019Leaderboards} &47.9 & 14.2&21.9 &11.3 &11.3 &10.7 &&&{NLP-TDMS-v2} & TDMS-IE&TA\textsuperscript{\#}\\

\citet{Kardas2020AXCELL} & 65.8 &58.5& 61.9& 56.0& 55.8 &54.1 &&& {NLP-TDMS-v3} &AxCell&\\
\citet{Kardas2020AXCELL} &53.4& 66.3& 59.2& 57.1 &66.1& 58.5 &&& {NLP-TDMS-v3}&TDMS-IE&\\
\citet{Kardas2020AXCELL} &67.8 &47.8& 56.1 &47.9& 46.4& 43.5 &&&{PwC-LB-v1}& AxCell&\\

\citet{Kabongo2021Mining} & 76.4&66.4&71.1&63.5&64.1&61.4&&& {NLP-TDMS-v1} &ORKG-TDM& XLNet \\ 
\citet{Kabongo2021Mining} &65.3&73.1&69.0&57.6&68.7&60.1&&& {NLP-TDMS-v1} &ORKG-TDM& SciBERT \\ 
\citet{Kabongo2021Mining} &79.5 &57.6&66.8&59.0&55.4&54.7&&& {NLP-TDMS-v1} &ORKG-TDM& BERT\\ 

\citet{Kabongo2021Mining} &77.1 &70.9&73.9&71.7&73.9& 70.6&&&{NLP-TDMS-v2} & ORKG-TDM&XLNet\\ 
\citet{Kabongo2021Mining} & 79.6 &63.3&70.5&68.1&67.5& 65.5&&&{NLP-TDMS-v2} & ORKG-TDM&BERT\\ 
\citet{Kabongo2021Mining} &65.7 &76.8&70.8&65.7&77.2&68.3&&&{NLP-TDMS-v2} & ORKG-TDM&SciBERT\\ 

\citet{Kabongo2021Mining} &95.1 &92 &93.5 & 92.3& 93.5& 91.7&&& {ORKG-PwC-v1}& ORKG-TDM&XLNet TAET\textsuperscript{\#}\\
\citet{Kabongo2021Mining} &93.5 & 93.2& 93.3&90.5 &94.4 &91.2 &&& {ORKG-PwC-v1}& ORKG-TDM&XLNet TAT\textsuperscript{\#}\\
\citet{Kabongo2021Mining} & 95.0& 90.5& 92.7& 91.6& 93.1& 91.2&&&{ORKG-PwC-v1}& ORKG-TDM&XLNet\textsuperscript{\#}\\
\citet{Kabongo2021Mining} & 95.7&88.3&91.8&91.7&92.1&90.8&&& {ORKG-PwC-v1} & ORKG-TDM&BERT\\
\citet{Kabongo2021Mining} &94.2 &89 &91.5 &89.2 &91.5&89.2 &&& {ORKG-PwC-v1}& ORKG-TDM&XLNet TAE\textsuperscript{\#}\\
\citet{Kabongo2021Mining} & 94.4&87.6&90.9&89.7&91.4&89.4&&& {ORKG-PwC-v1} & ORKG-TDM&SciBERT \\
\citet{Kabongo2021Mining} & 92.6 &90 &91.3&88.6&92.9&89.4 &&& {ORKG-PwC-v1}& ORKG-TDM&XLNet TA\textsuperscript{\#} \\

\citet{Kabongo2021Mining} &94.9&91.2&93.0&92.8&94.8&92.8&&& {ORKG-PwC-v2} & ORKG-TDM&XLNet\\
\citet{Kabongo2021Mining} & 95.5 & 89.1 & 92.1 & 92.8 & 93.9 & 92.4 &&& {ORKG-PwC-v2} & ORKG-TDM&BERT\\
\citet{Kabongo2021Mining} & 94.1&88.5&91.2&90.9&93.4&91.1&&& {ORKG-PwC-v2} & ORKG-TDM&SciBERT\\

\citet{Kabongo2023ORKG} &95.2&92.2&93.6&91.5&93.3&91.3 &&& {ORKG-PwC-v5} & ORKG-LB&BigBERT\\ 
\citet{Kabongo2023ORKG} & 94.8&93.9&94.3&91.3&94.4&91.8 &&& {ORKG-PwC-v5} & ORKG-LB&BERT\\
\citet{Kabongo2023ORKG} & 94.8&93.9&94.3&91.3&94.4&91.8 &&& {ORKG-PwC-v5} & ORKG-LB&SciBERT\\

\citet{Kabongo2023ORKG} & 95.4&93.9&94.7&93.2&95.7&93.5&&& {ORKG-PwC-v6} & ORKG-LB&BERT\\
\citet{Kabongo2023ORKG} & 95.4 & 91.1 & 93.2 & 92.6 & 94.3 & 92.2 &&& {ORKG-PwC-v6} & ORKG-LB&SciBERT \\
\citet{Kabongo2023ORKG} & 93.2&94.9&93.0&95.7&92.4&94.0 &&& {ORKG-PwC-v6} & ORKG-LB&BigBERT \\
\citet{Kabongo2023ORKG}& 95.1 & 94.6 & 94.8 & 93.1 & 96.4 & 93.7 &&& {ORKG-PwC-v6} & ORKG-LB&XLNet\\

\citet{Kabongo2023ORKG} & 95.4 & 88.0 & 91.5 & 91.2 & 92.3 & 90.6 &&& {ORKG-PwC-v3} & ORKG-LB&BERT \\
\citet{Kabongo2023ORKG} & 93.7 & 86.0 & 89.7 & 89.4 & 91.7 & 89.2 &&& {ORKG-PwC-v3} & ORKG-LB&SciBERT\\
\citet{Kabongo2023ORKG}& 93.6 & 85.3 & 89.3 & 87.5 & 88.7 & 86.6 &&& {ORKG-PwC-v3} & ORKG-LB&BigBird\\

\citet{Kabongo2023ORKG} & 94.9 & 91.2 & 93.0 & 91.9 & 94.4 & 92.0 &&& {ORKG-PwC-v4} & ORKG-LB&XLNet\\
\citet{Kabongo2023ORKG}& 96.0 & 90.0 & 92.9 & 93.5 & 94.2 & 92.8 &&& {ORKG-PwC-v4}& ORKG-LB&BERT\\
\citet{Kabongo2023ORKG}& 94.6 & 88.6 & 91.5 & 91.7 & 93.9 & 91.6 &&& {ORKG-PwC-v4} & ORKG-LB&SciBERT\\
\citet{Kabongo2023ORKG} & 94.6 & 87.2 & 90.7 & 90.7 & 91.6 & 89.7 &&& {ORKG-PwC-v4} & ORKG-LB&BigBird\\

\citet{Kabongo2023ZeroShot}&9.2&78.1&16.5&14.3&86.6&21.9&&&{ORKG-PwC-v7\textsuperscript{*}}&ORKG-TDM&XLNet\\
\citet{Kabongo2023ZeroShot} & 14.1&72.9&23.6&20.1&83.4&28.9&&&{ORKG-PwC-v7\textsuperscript{*}}& ORKG-TDM&BERT\\

\citet{Kabongo2023ZeroShot}& 10.4&81.7&18.4&16.2&89&24.4&&&{ORKG-PwC-v7\textsuperscript{*}}& ORKG-TDM&BERT\\
\citet{Kabongo2023ZeroShot}& 10.1&76.8&17.8&14.9&86.4&22.7 &&&{ORKG-PwC-v7\textsuperscript{*}}&ORKG-TDM&XLNet\\

\citet{sahinuc2024efficient}&55.1&25.8&35.1&&&&&&SciLead& AxCell & \\ 
\citet{sahinuc2024efficient}&40.7&39.5&40.1&&&&&&SciLead& TDMR-PR &Llama 2+CS \\ 
\citet{sahinuc2024efficient}&35.9&34.9&35.4&&&&&&SciLead& TDMR-PR &Llama 2 \\ 
\citet{sahinuc2024efficient}&58.4&52.1&55.1&&&&&&SciLead& TDMR-PR &Mixtral+CS \\ 
\citet{sahinuc2024efficient}&55.7&48.8&51.0&&&&&&SciLead& TDMR-PR &Mixtral \\ 
\citet{sahinuc2024efficient}&62.0&58.1&60.0&&&&&&SciLead& TDMR-PR &Llama 3+CS \\ 
\citet{sahinuc2024efficient}&77.1&72.6&74.8&&&&&&SciLead& TDMR-PR &Llama 3 \\ 
\citet{sahinuc2024efficient}&69.0&63.8&66.3&&&&&&SciLead& TDMR-PR &GPT-4+CS \\ 
\citet{sahinuc2024efficient}&75.3&70.4&72.8&&&&&&SciLead& TDMR-PR &GPT-4 \\ 

\bottomrule
 \multicolumn{12}{l}{\rule{0pt}{1.5ex} \textbf{Open Domain Problem Framing}}\\[0.1ex]
\toprule

\citet{Yang2022TELIN}& 68.2 &45.3 &56.5 &49.7 &43.1 &42.5&&&{PwC-LB-v2}& TELIN&\\ 

\bottomrule
 \multicolumn{12}{l}{\rule{0pt}{1.5ex} \textbf{Hybrid Domain Problem Framing}}\\[0.1ex]
\toprule

\citet{sahinuc2024efficient}&27.23&22.99&24.93&&&&&&SciLead& TDMR-PR & Llama 2\\
\citet{sahinuc2024efficient}&27.89&24.48&26.07&&&&&&SciLead& TDMR-PR & Mixtral \\

\citet{sahinuc2024efficient}&50.75&45.30&47.87&&&&&&SciLead& TDMR-PR & Llama 3 \\
\citet{sahinuc2024efficient}&56.08&51.89&53.90&&&&&&SciLead& TDMR-PR & GPT-4 \\ 
\bottomrule

\multicolumn{12}{p{16cm}}{\textsuperscript{\textdagger}SM, MLC, and EL are baseline methods, representing String Match, Multi-Label Classification, and Entity Linking, respectively. 
\textsuperscript{*} trained on ORKG-PwC-v6/v7. \textsuperscript{\#} REC, TAET, and Full refer to DocREC, DocTAET, and the Full Paper representations of the document, respectively. These are reported as part of an ablation study examining different document representations. For more details on these representations, see \autoref{sec:doc_rep}.}

\end{tabular}

\caption{\fontsize{10pt}{12pt}\selectfont Summary of results for leaderboard \textbf{$\langle$Task, Dataset, Metric$\rangle$} extraction.\label{tab:task-dataset-metric}}
\end{table*}

\begin{table*}

\setlength{\tabcolsep}{1.0pt}
\fontsize{8pt}{10pt}\selectfont 

\renewcommand{\arraystretch}{0.9}
\begin{tabular}{l N N N N N N N N L{2.2cm} L{2cm} p{2.8cm}}
\toprule
&\multicolumn{3}{c}{\textbf{Micro}} & \multicolumn{3}{c}{\textbf{Macro}}& \multicolumn{2}{c}{\textbf{Part. Micro}} &&\\
\cmidrule(lr){2-4} \cmidrule(lr){5-7} \cmidrule(lr){8-9} 
\textbf{Reported In}& \textbf{P} &\textbf{R} & \textbf{F1} & \textbf{P} & \textbf{R} & \textbf{F1} & \textbf{P} & \textbf{F1} & \textbf{Dataset} & \textbf{Method}& \textbf{Experimental Setup}\\

 \bottomrule
 \multicolumn{12}{l}{\rule{0pt}{1.5ex} \textbf{Closed Domain Problem Framing}}\\[0.1ex]
\toprule

\citet{Hou2019Leaderboards}& 10.8& 13.1 & 11.8 & 9.3 & 11.8 & 9.9 &&& {NLP-TDMS-v1} & TDMS-IE&\\
\citet{Hou2019Leaderboards}&3.8&1.8&2.4&1.3&1.0&1.1 &&& {NLP-TDMS-v1} &SM{\textsuperscript{\textdagger}}&\\
\citet{Hou2019Leaderboards}& 6.8& 2.9& 4.0& 6.8&6.1&6.2 &&&{NLP-TDMS-v1} &MLC{\textsuperscript{\textdagger}}&\\
\citet{Kardas2020AXCELL} & 27.4 &24.4& 25.8& 20.2& 20.6& 19.7 &&& {NLP-TDMS-v3} &AxCell&\\
\citet{Kardas2020AXCELL}& 6.8 &8.4& 7.5& 8.6& 9.5& 8.8 &&& {NLP-TDMS-v3} &TDMS-IE &\\

\citet{Kardas2020AXCELL} & 37.4& 23.2& 28.7 &24.0& 21.8& 21.1 &&&{PwC-LB-v1}& AxCell&\\

\citet{sahinuc2024efficient}&32.59&13.67&19.26&&&&&&SciLead& AxCell &\\ 
\citet{sahinuc2024efficient}&10.06&21.59&13.73&&&&&&SciLead& TDMR-PR &Llama 2+CS \\ 
\citet{sahinuc2024efficient}&9.63&15.25&11.81&&&&&&SciLead& TDMR-PR &Llama 2 \\ 
\citet{sahinuc2024efficient}&26.54&24.61&25.54&&&&&&SciLead& TDMR-PR &Mixtral+CS \\ 
\citet{sahinuc2024efficient}&24.66&21.73&23.10&&&&&&SciLead& TDMR-PR &Mixtral \\ 
\citet{sahinuc2024efficient}&23.22&29.54&26.00&&&&&&SciLead& TDMR-PR &Llama 3+CS \\ 
\citet{sahinuc2024efficient}&27.11&35.60&30.78&&&&&&SciLead& TDMR-PR &Llama 3 \\ 
\citet{sahinuc2024efficient}&49.82&48.71&49.26&&&&&&SciLead& TDMR-PR &GPT-4+CS \\ 
\citet{sahinuc2024efficient}&56.02&54.53&55.27&&&&&&SciLead& TDMR-PR &GPT-4 \\ 

 \bottomrule
 \multicolumn{12}{l}{\rule{0pt}{1.5ex} \textbf{Open Domain Problem Framing}}\\[0.1ex]
\toprule

\citet{Yang2022TELIN}&38.3 &20.8 &26.3& 26.6 &19.2& 21.3 &&&{PwC-LB-v2}& TELIN\\

 \bottomrule
 \multicolumn{12}{l}{\rule{0pt}{1.5ex} \textbf{Hybrid Domain Problem Framing}}\\[0.1ex]
\toprule

\citet{sahinuc2024efficient}&4.17&9.89&5.87&&&&&&SciLead& TDMR-PR & Llama 2\\ 
\citet{sahinuc2024efficient}&14.65&12.27&13.35&&&&&&SciLead& TDMR-PR & Mixtral \\ 
\citet{sahinuc2024efficient}&15.70&18.75&17.09&&&&&&SciLead& TDMR-PR & Llama 3 \\ 
\citet{sahinuc2024efficient}&40.60&39.56&40.07&&&&&&SciLead& TDMR-PR & GPT-4 \\ 

\citet{sahinuc2024efficient}&51.01&51.03&51.02&&&&&&SciLead& TDMR-PR & GPT-4 FS\\ 

\bottomrule

\multicolumn{12}{p{16cm}}{\textsuperscript{\textdagger}SM, MLC, and EL are baseline methods, representing String Match, Multi-Label Classification, and Entity Linking, respectively.}
\end{tabular}
\caption{\fontsize{10pt}{12pt}\selectfont Summary of results for leaderboard \textbf{$\langle$Task, Dataset, Metric, Score$\rangle$} extraction. Notations: \textbf{FS} = Few Shot.}

\end{table*}

\begin{table*}

\setlength{\tabcolsep}{1.0pt}
\fontsize{8pt}{10pt}\selectfont 

\renewcommand{\arraystretch}{0.9}
\begin{tabular}{l N N N N N N N N L{2.2cm} L{2cm} p{2.8cm}}
\toprule
&\multicolumn{3}{c}{\textbf{Micro}} & \multicolumn{3}{c}{\textbf{Macro}}& \multicolumn{2}{c}{\textbf{Part. Micro}} &&\\
\cmidrule(lr){2-4} \cmidrule(lr){5-7} \cmidrule(lr){8-9} 
\textbf{Reported In}& \textbf{P} &\textbf{R} & \textbf{F1} & \textbf{P} & \textbf{R} & \textbf{F1} & \textbf{P} & \textbf{F1} & \textbf{Dataset} & \textbf{Method}& \textbf{Experimental Setup}\\

 \bottomrule

 \multicolumn{12}{l}{\rule{0pt}{1.5ex} \textbf{Closed Domain Problem Framing}}\\[0.1ex]
\toprule

\citet{Jain2020SCIREX}&0.48 & 0.89& 0.62&&&&&&{SciREX}&TDMS-IE &\\

\bottomrule
\multicolumn{12}{l}{\rule{0pt}{1.5ex} \textbf{Open Domain Problem Framing}}\\[0.1ex]
\toprule

\citet{Jain2020SCIREX}&0.53 & 0.72 & 0.61 &&&&&& {SciREX}&SciREX-IE&\\

\bottomrule

\end{tabular}
\caption{\fontsize{10pt}{12pt}\selectfont Summary of results for leaderboard \textbf{$\langle$Task, Dataset, Metric, Method$\rangle$} extraction.}
\end{table*}

\begin{table*}

\setlength{\tabcolsep}{1.0pt}
\fontsize{8pt}{10pt}\selectfont 

\renewcommand{\arraystretch}{0.9}
\begin{tabular}{l N N N N N N N N L{2.2cm} L{2cm} p{2.8cm}}
\toprule
&\multicolumn{3}{c}{\textbf{Micro}} & \multicolumn{3}{c}{\textbf{Macro}}& \multicolumn{2}{c}{\textbf{Part. Micro}} &&\\
\cmidrule(lr){2-4} \cmidrule(lr){5-7} \cmidrule(lr){8-9} 
\textbf{Reported In}& \textbf{P} &\textbf{R} & \textbf{F1} & \textbf{P} & \textbf{R} & \textbf{F1} & \textbf{P} & \textbf{F1} & \textbf{Dataset} & \textbf{Method}& \textbf{Experimental Setup}\\
\bottomrule

\multicolumn{12}{l}{\rule{0pt}{1.5ex} \textbf{Closed Domain Problem Framing}}\\[0.1ex]
\toprule

\citet{Kardas2020AXCELL} & 70.6 &57.3 &63.3& 60.7& 62.6 &59.7 & &&{PwC-LB-v1}& AxCell&\\
\citet{Kabongo2021Mining} & 97.4& 93.6& 95.5& 93.7& 94.8& 93.6& & & {ORKG-PwC-v1}& ORKG-TDM &XLNet\\
\citet{Kabongo2023ORKG} & 96.8 & 95.9 & 96.4 & 94.3 & 97.2 & 95.0 & && {ORKG-PwC-v6} & ORKG-LB &XLNet\\
\citet{Kabongo2023ORKG} & 96.8 & 95.9 & 96.4 & 94.3 & 97.2 & 95.0 & &&{ORKG-PwC-v4} & ORKG-LB &XLNet\\

\citet{sahinuc2024efficient}&68.98&58.52&63.32&&&&&&SciLead& AxCell &\\ 
\citet{sahinuc2024efficient}&59.83&67.20&63.30&&&&&&SciLead& TDMR-PR &Llama 2+CS \\ 
\citet{sahinuc2024efficient}&55.45&60.74&57.97&&&&&&SciLead& TDMR-PR &Llama 2 \\ 
\citet{sahinuc2024efficient}&86.27&91.99&89.04&&&&&&SciLead& TDMR-PR &Mixtral+CS \\ 
\citet{sahinuc2024efficient}&86.85&89.74&88.27&&&&&&SciLead& TDMR-PR &Mixtral \\ 
\citet{sahinuc2024efficient}&85.69&90.85&88.19&&&&&&SciLead& TDMR-PR &Llama 3+CS \\ 
\citet{sahinuc2024efficient}&87.33&92.17&89.68&&&&&&SciLead& TDMR-PR &Llama 3 \\ 
\citet{sahinuc2024efficient}&90.70&90.77&90.73&&&&&&SciLead& TDMR-PR &GPT-4+CS \\ 
\citet{sahinuc2024efficient}&90.62&91.10&90.86&&&&&&SciLead& TDMR-PR &GPT-4 \\ 

\bottomrule
\multicolumn{12}{l}{\rule{0pt}{1.5ex} \textbf{Open Domain Problem Framing}}\\[0.1ex]
\toprule

\citet{Yang2022TELIN}&70.3 &53.7 &59.2 &60.5 &57.3 &57.1 & & &{PwC-LB-v2}& TELIN&\\

\citet{kabongo2024effective}&31.89&&13.97&&&&54.92&24.05&{TDMS-Ctx-v5}& TDMS-PR& Llama2 7B ZS REC\textsuperscript{\#}\\
\citet{kabongo2024effective}&24.56&&21.75&&&&43.46&38.48&{TDMS-Ctx-v6}& TDMS-PR& Llama2 7B ZS TAET\textsuperscript{\#}\\
\citet{kabongo2024effective}&2.06&&2.06&&&&52.54&3.36&{TDMS-Ctx-v4}& TDMS-PR& Llama2 7B ZS Full\textsuperscript{\#}\\
\citet{kabongo2024effective}&17.99&&17.99&&&&59.25&29.88&{TDMS-Ctx-v5}& TDMS-PR& Mistral 7B ZS REC\textsuperscript{\#}\\
\citet{kabongo2024effective}&26.99&&26.99&&&&64.00&44.90&{TDMS-Ctx-v6}& TDMS-PR& Mistral 7B ZS TAET\textsuperscript{\#}\\
\citet{kabongo2024effective}&0.22&&0.56&&&&62.50&0.56&{TDMS-Ctx-v4}& TDMS-PR& Mistral 7B ZS Full\textsuperscript{\#}\\

\citet{kabongo2024effective}&34.10&&20.93&&&&51.13&31.37&{TDMS-Ctx-v2}& TDMS-PR& Llama2 7B FS REC\textsuperscript{\#}\\
\citet{kabongo2024effective}&30.61&&29.53&&&&44.96&43.37&{TDMS-Ctx-v3}& TDMS-PR& Llama2 7B FS TAET\textsuperscript{\#} \\
\citet{kabongo2024effective}&34.69&&1.59&&&&50.00&2.29&{TDMS-Ctx-v1}& TDMS-PR& Llama2 7B FS Full\textsuperscript{\#}\\
\citet{kabongo2024effective}&37.65&&26.77&&&&55.90&39.75&{TDMS-Ctx-v2}& TDMS-PR& Mistral 7B FS REC\textsuperscript{\#}\\
\citet{kabongo2024effective}&39.48&&33.38&&&&54.82&46.35&{TDMS-Ctx-v3}& TDMS-PR& Mistral 7B FS TAET\textsuperscript{\#} \\
\citet{kabongo2024effective}&32.43&&0.81&&&&71.43&1.19&{TDMS-Ctx-v1}& TDMS-PR& Mistral 7B FS Full\textsuperscript{\#}\\

\bottomrule
\multicolumn{12}{l}{\rule{0pt}{1.5ex} \textbf{Hybrid Domain Problem Framing}}\\[0.1ex]
\toprule

\citet{sahinuc2024efficient}&39.70&42.98&41.27&&&&&&SciLead& TDMR-PR & Llama 2\\ 
\citet{sahinuc2024efficient}&50.23&60.72&54.98&&&&&&SciLead& TDMR-PR & Mixtral\\ 
\citet{sahinuc2024efficient}&65.72&80.39&72.32&&&&&&SciLead& TDMR-PR & Llama 3\\ 
\citet{sahinuc2024efficient}&63.82&78.30&70.32&&&&&&SciLead& TDMR-PR & GPT-4\\ 

\bottomrule

\multicolumn{12}{p{16cm}}{\textsuperscript{\#} REC, TAET, and Full refer to DocREC, DocTAET, and the Full Paper representations of the document, respectively. These are reported as part of an ablation study examining different document representations. For more details on these representations, see \autoref{sec:doc_rep}.}\\

\end{tabular}
\caption{\fontsize{10pt}{12pt}\selectfont Summary of results for leaderboard \textbf{$\langle$Task$\rangle$} extraction. Notations: \textbf{FS} = Few Shot, \textbf{ZS} = Zero Shot.}
\end{table*}

\begin{table*}

\setlength{\tabcolsep}{1.0pt}
\fontsize{8pt}{10pt}\selectfont 

\renewcommand{\arraystretch}{0.9}
\begin{tabular}{l N N N N N N N N L{2.2cm} L{2cm} p{2.8cm}}
\toprule
&\multicolumn{3}{c}{\textbf{Micro}} & \multicolumn{3}{c}{\textbf{Macro}}& \multicolumn{2}{c}{\textbf{Part. Micro}} &&\\
\cmidrule(lr){2-4} \cmidrule(lr){5-7} \cmidrule(lr){8-9} 
\textbf{Reported In}& \textbf{P} &\textbf{R} & \textbf{F1} & \textbf{P} & \textbf{R} & \textbf{F1} & \textbf{P} & \textbf{F1} & \textbf{Dataset} & \textbf{Method}& \textbf{Experimental Setup}\\

 \bottomrule

\multicolumn{12}{l}{\rule{0pt}{1.5ex} \textbf{Results of Extracting \(\langle\)Dataset\(\rangle\) for Closed Domain Problem Framing}}\\[0.1ex]
\toprule

\citet{Kardas2020AXCELL} & 70.2 &48.4& 57.3& 53.5& 52.7 &49.9 & & &{PwC-LB-v1}& AxCell\\
\citet{Kabongo2021Mining} & 96.6& 91.5& 94.0& 92.9& 93.6& 92.4& & &{ORKG-PwC-v1}& ORKG-TDM &XLNet\\
\citet{Kabongo2023ORKG}& 96.2 & 95.4 & 95.8 & 93.8 & 96.7 & 94.4 && &{ORKG-PwC-v6} & ORKG-LB &XLNet \\
\citet{Kabongo2023ORKG} & 96.2 & 95.4 & 95.8 & 93.8 & 96.7 & 94.4 & && {ORKG-PwC-v4} & ORKG-LB &XLNet \\

\citet{sahinuc2024efficient}&63.66&33.87&44.22&&&&&&SciLead& AxCell \\ 
\citet{sahinuc2024efficient}&68.93&58.81&63.47&&&&&&SciLead& TDMR-PR &Llama 2+CS \\ 
\citet{sahinuc2024efficient}&62.60&55.03&58.57&&&&&&SciLead& TDMR-PR &Llama 2 \\ 
\citet{sahinuc2024efficient}&85.03&73.20&78.67&&&&&&SciLead& TDMR-PR &Mixtral+CS \\ 
\citet{sahinuc2024efficient}&81.68&71.26&76.12&&&&&&SciLead& TDMR-PR &Mixtral \\ 
\citet{sahinuc2024efficient}&82.43&78.62&80.48&&&&&&SciLead& TDMR-PR &Llama 3+CS \\ 
\citet{sahinuc2024efficient}&92.09&87.75&89.87&&&&&&SciLead& TDMR-PR &Llama 3 \\ 
\citet{sahinuc2024efficient}&86.36&79.93&83.02&&&&&&SciLead& TDMR-PR &GPT-4+CS \\ 
\citet{sahinuc2024efficient}&92.64&86.05&89.22&&&&&&SciLead& TDMR-PR &GPT-4 \\ 

\bottomrule
\multicolumn{12}{l}{\rule{0pt}{1.5ex} \textbf{Results of Extracting \(\langle\)Dataset\(\rangle\) for Open Domain Problem Framing}}\\[0.1ex]
\toprule

\citet{kabongo2024effective}&15.77&&6.83&&&&38.32&16.6&{TDMS-Ctx-v5}& TDMS-PR& Llama2 7B ZS REC\textsuperscript{\#}\\
\citet{kabongo2024effective}&12.72&&11.26&&&&26.09&23.1&{TDMS-Ctx-v6}& TDMS-PR& Llama2 7B ZS TAET\textsuperscript{\#} \\
\citet{kabongo2024effective}&20.34&&1.30&&&&38.98&2.49&{TDMS-Ctx-v4}& TDMS-PR& Llama2 7B ZS Full\textsuperscript{\#}\\
\citet{kabongo2024effective}&23.40&&11.80&&&&41.73&21.05&{TDMS-Ctx-v5}& TDMS-PR& Mistral 7B ZS REC\textsuperscript{\#}\\
\citet{kabongo2024effective}&20.41&&14.32&&&&38.89&27.29&{TDMS-Ctx-v6}& TDMS-PR& Mistral 7B ZS TAET\textsuperscript{\#} \\
\citet{kabongo2024effective}&37.50&&0.33&&&&75.00&0.67&{TDMS-Ctx-v4}& TDMS-PR& Mistral 7B ZS Full\textsuperscript{\#}\\

\citet{kabongo2024effective}&21.27&&13.06&&&&36.66&22.50&{TDMS-Ctx-v2}& TDMS-PR& Llama2 7B FS REC\textsuperscript{\#}\\
\citet{kabongo2024effective}&17.29&&16.68&&&&31.48&30.36&{TDMS-Ctx-v3}& TDMS-PR& Llama2 7B FS TAET\textsuperscript{\#} \\
\citet{kabongo2024effective}&29.59&&1.36&&&&39.80&1.82&{TDMS-Ctx-v1}& TDMS-PR& Llama2 7B FS Full\textsuperscript{\#}\\
\citet{kabongo2024effective}&22.15&&15.68&&&&38.52&27.28&{TDMS-Ctx-v2}& TDMS-PR& Mistral 7B FS REC\textsuperscript{\#}\\
\citet{kabongo2024effective}&21.89&&18.51&&&&38.73&32.75&{TDMS-Ctx-v3}& TDMS-PR& Mistral 7B FS TAET\textsuperscript{\#}\\
\citet{kabongo2024effective}&32.43&&0.57&&&&48.65&0.85&{TDMS-Ctx-v1}& TDMS-PR& Mistral 7B FS Full\textsuperscript{\#}\\

\citet{Yang2022TELIN}&70.9& 52.8& 59.3& 54.7& 55.2 &53.9& & & {PwC-LB-v2}& TELIN&\\

\bottomrule
\multicolumn{12}{l}{\rule{0pt}{1.5ex} \textbf{Results of Extracting \(\langle\)Dataset\(\rangle\) for Hybrid Domain Problem Framing}}\\[0.1ex]
\toprule

\citet{sahinuc2024efficient}&41.05&33.14&36.67&&&&&&SciLead& TDMR-PR & Llama 2\\ 
\citet{sahinuc2024efficient}&49.67&44.45&46.92&&&&&&SciLead& TDMR-PR & Mixtral\\ 
\citet{sahinuc2024efficient}&66.81&62.86&64.77&&&&&&SciLead& TDMR-PR & Llama 3\\ 
\citet{sahinuc2024efficient}&83.29&79.52&81.36&&&&&&SciLead& TDMR-PR & GPT-4\\ 

\bottomrule

\multicolumn{12}{p{16cm}}{\textsuperscript{\#} REC, TAET, and Full refer to DocREC, DocTAET, and the Full Paper representations of the document, respectively. These are reported as part of an ablation study examining different document representations. For more details on these representations, see \autoref{sec:doc_rep}.}\\
\end{tabular}
\caption{\fontsize{10pt}{12pt}\selectfont Summary of results for leaderboard \textbf{$\langle$Dataset$\rangle$} extraction. Notations: \textbf{FS} = Few Shot, \textbf{ZS} = Zero Shot.}
\end{table*}

\begin{table*}

\setlength{\tabcolsep}{1.0pt}
\fontsize{8pt}{10pt}\selectfont 

\renewcommand{\arraystretch}{0.9}
\begin{tabular}{l N N N N N N N N L{2.2cm} L{2cm} p{2.8cm}}
\toprule
&\multicolumn{3}{c}{\textbf{Micro}} & \multicolumn{3}{c}{\textbf{Macro}}& \multicolumn{2}{c}{\textbf{Part. Micro}} &&\\
\cmidrule(lr){2-4} \cmidrule(lr){5-7} \cmidrule(lr){8-9} 
\textbf{Reported In}& \textbf{P} &\textbf{R} & \textbf{F1} & \textbf{P} & \textbf{R} & \textbf{F1} & \textbf{P} & \textbf{F1} & \textbf{Dataset} & \textbf{Method}& \textbf{Experimental Setup}\\

 \bottomrule

\multicolumn{12}{l}{\rule{0pt}{1.5ex} \textbf{Closed Domain Problem Framing}}\\[0.1ex]
\toprule
 
\citet{Kardas2020AXCELL} & 68.8 &58.5& 63.3& 58.4& 60.4& 56.5 & &&{PwC-LB-v1}& AxCell\\
\citet{Kabongo2021Mining}& 96.0& 92.5& 94.2& 92.5& 94.2& 92.5& &&{ORKG-PwC-v1}& ORKG-TDM &XLNet\\
\citet{Kabongo2023ORKG} & 96.0 & 95.3 & 95.6 & 93.7 & 96.9 & 94.4 & &&{ORKG-PwC-v6} & ORKG-LB &XLNet\\
\citet{Kabongo2023ORKG}& 96.0 & 95.3 & 95.6 & 93.7 & 96.9 & 94.4 & &&{ORKG-PwC-v4} & ORKG-LB &XLNet\\

 \citet{kabongo2024effective}&26.77&&11.72&&&&41.73&18.28&{TDMS-Ctx-v5}& TDMS-PR& Llama2 7B ZS REC\textsuperscript{\#}\\
\citet{kabongo2024effective}&19.19&&16.99&&&&30.60&27.09&{TDMS-Ctx-v6}& TDMS-PR& Llama2 7B ZS TAET\textsuperscript{\#}\\
\citet{kabongo2024effective}&23.73&&1.52&&&&38.98&2.49&{TDMS-Ctx-v4}& TDMS-PR& Llama2 7B ZS Full\textsuperscript{\#}\\
\citet{kabongo2024effective}&31.02&&15.55&&&&46.20&23.16&{TDMS-Ctx-v5}& TDMS-PR& Mistral 7B ZS REC\textsuperscript{\#}\\
\citet{kabongo2024effective}&31.41&&22.04&&&&45.94&32.23&{TDMS-Ctx-v6}& TDMS-PR& Mistral 7B ZS TAET\textsuperscript{\#}\\
\citet{kabongo2024effective}&37.50&&0.33&&&&87.50&0.78&{TDMS-Ctx-v4}& TDMS-PR& Mistral 7B ZS Full\textsuperscript{\#}\\

\citet{kabongo2024effective}&22.74&&13.96&&&&35.82&21.99&{TDMS-Ctx-v2}& TDMS-PR& Llama2 7B FS REC\textsuperscript{\#}\\
\citet{kabongo2024effective}&20.78&&20.02&&&&31.66&30.51&{TDMS-Ctx-v3}& TDMS-PR & Llama2 7B FS TAET\textsuperscript{\#}\\
\citet{kabongo2024effective}&20.41&&0.94&&&&36.73&1.68&{TDMS-Ctx-v1}& TDMS-PR & Llama2 7B FS Full\textsuperscript{\#}\\
\citet{kabongo2024effective}&26.38&&18.70&&&&40.18&28.49&{TDMS-Ctx-v2}& TDMS-PR & Mistral 7B FS REC\textsuperscript{\#}\\
\citet{kabongo2024effective}&28.66&&24.23&&&&40.41&34.16&{TDMS-Ctx-v3}& TDMS-PR & Mistral 7B FS TAET\textsuperscript{\#}\\
\citet{kabongo2024effective}&32.43&&0.57&&&&45.95&0.81&{TDMS-Ctx-v1}& TDMS-PR & Mistral 7B FS Full\textsuperscript{\#}\\

\citet{sahinuc2024efficient}&69.35&51.36&59.01&&&&&&SciLead& AxCell \\ 
\citet{sahinuc2024efficient}&67.36&61.41&64.25&&&&&&SciLead& TDMR-PR &Llama 2+CS \\ 
\citet{sahinuc2024efficient}&71.51&65.49&68.37&&&&&&SciLead& TDMR-PR &Llama 2 \\ 
\citet{sahinuc2024efficient}&76.56&71.78&74.09&&&&&&SciLead& TDMR-PR &Mixtral+CS \\ 
\citet{sahinuc2024efficient}&76.72&67.20&71.65&&&&&&SciLead& TDMR-PR &Mixtral \\ 
\citet{sahinuc2024efficient}&87.02&81.41&84.12&&&&&&SciLead& TDMR-PR &Llama 3+CS \\ 
\citet{sahinuc2024efficient}&94.90&89.48&92.11&&&&&&SciLead& TDMR-PR &Llama 3 \\ 
\citet{sahinuc2024efficient}&86.36&81.49&83.85&&&&&&SciLead& TDMR-PR &GPT-4+CS \\ 
\citet{sahinuc2024efficient}&88.18&86.46&87.31&&&&&&SciLead& TDMR-PR &GPT-4 \\ 

\bottomrule
\multicolumn{12}{l}{\rule{0pt}{1.5ex} \textbf{Open Domain Problem Framing}}\\[0.1ex]
\toprule

\citet{Yang2022TELIN}&63.2 &57.9& 60.2 &56.3& 55.1 &55.4& &&{PwC-LB-v2}& TELIN\\

\bottomrule
\multicolumn{12}{l}{\rule{0pt}{1.5ex} \textbf{Hybrid Domain Problem Framing}}\\[0.1ex]
\toprule

\citet{sahinuc2024efficient}&61.24&59.34&60.28&&&&&&SciLead& TDMR-PR & Llama 2\\ 
\citet{sahinuc2024efficient}&78.72&71.19&74.77&&&&&&SciLead& TDMR-PR & Mixtral \\ 
\citet{sahinuc2024efficient}&94.90&88.90&91.80&&&&&&SciLead& TDMR-PR & Llama 3 \\ 
\citet{sahinuc2024efficient}&92.21&89.27&90.72&&&&&&SciLead& TDMR-PR & GPT-4 \\ 

\bottomrule
\multicolumn{12}{p{16cm}}{\textsuperscript{\#} REC, TAET, and Full refer to DocREC, DocTAET, and the Full Paper representations of the document, respectively. These are reported as part of an ablation study examining different document representations. For more details on these representations, see \autoref{sec:doc_rep}.}\\

\end{tabular}
\caption{\fontsize{10pt}{12pt}\selectfont Summary of results for leaderboard \textbf{$\langle$Metric$\rangle$} extraction.Notations: \textbf{FS} = Few Shot, \textbf{ZS} = Zero Shot.}
\end{table*}

\begin{table*}

\setlength{\tabcolsep}{1.0pt}
\fontsize{8pt}{10pt}\selectfont 

\renewcommand{\arraystretch}{0.9}
\begin{tabular}{l N N N N N N N N L{2.2cm} L{2cm} p{2.8cm}}
\toprule
&\multicolumn{3}{c}{\textbf{Micro}} & \multicolumn{3}{c}{\textbf{Macro}}& \multicolumn{2}{c}{\textbf{Part. Micro}} &&\\
\cmidrule(lr){2-4} \cmidrule(lr){5-7} \cmidrule(lr){8-9} 
\textbf{Reported In}& \textbf{P} &\textbf{R} & \textbf{F1} & \textbf{P} & \textbf{R} & \textbf{F1} & \textbf{P} & \textbf{F1} & \textbf{Dataset} & \textbf{Method}& \textbf{Experimental Setup}\\

 \bottomrule

\multicolumn{12}{l}{\rule{0pt}{1.5ex} \textbf{Closed Domain Problem Framing}}\\[0.1ex]
\toprule

\citet{sahinuc2024efficient}&45.32&18.41&26.18&&&&&&SciLead& AxCell &\\ 
\citet{sahinuc2024efficient}&23.75&31.61&27.12&&&&&&SciLead& TDMR-PR &Llama 2 \\ 
\citet{sahinuc2024efficient}&44.62&41.75&43.13&&&&&&SciLead& TDMR-PR &Mixtral \\ 
\citet{sahinuc2024efficient}&39.50&49.56&43.96&&&&&&SciLead& TDMR-PR &Llama 3 \\ 
\citet{sahinuc2024efficient}&70.34&68.22&69.26&&&&&&SciLead& TDMR-PR &GPT-4 \\ 
\bottomrule

\multicolumn{12}{l}{\rule{0pt}{1.5ex} \textbf{Open Domain Problem Framing}}\\[0.1ex]
\toprule
 
\citet{kabongo2024effective}&6.06&&2.61&&&&7.27&3.10&{TDMS-Ctx-v5}& TDMS-PR& Llama2 7B ZS REC\textsuperscript{\#}\\
\citet{kabongo2024effective}&0.87&&0.77&&&&1.09&0.96&{TDMS-Ctx-v6}& TDMS-PR& Llama2 7B ZS TAET\textsuperscript{\#}\\
\citet{kabongo2024effective}&5.08&&0.33&&&&8.47&0.54&{TDMS-Ctx-v4}& TDMS-PR& Llama2 7B ZS Full\textsuperscript{\#}\\
\citet{kabongo2024effective}&9.98&&5.04&&&&11.46&5.75&{TDMS-Ctx-v5}& TDMS-PR& Mistral 7B ZS REC\textsuperscript{\#}\\
\citet{kabongo2024effective}&1.71&&1.20&&&&2.03&1.41&{TDMS-Ctx-v6}& TDMS-PR& Mistral 7B ZS TAET\textsuperscript{\#}\\
\citet{kabongo2024effective}&14.00&&0.76&&&&21.62&0.87&{TDMS-Ctx-v4}& TDMS-PR& Mistral 7B ZS Full\textsuperscript{\#}\\

\citet{kabongo2024effective}&4.99&&3.04&&&&5.59&3.46&{TDMS-Ctx-v2}& TDMS-PR& Llama2 7B FS REC\textsuperscript{\#}\\
\citet{kabongo2024effective}&1.18&&1.14&&&&1.43&1.38&{TDMS-Ctx-v3}& TDMS-PR& Llama2 7B FS TAET\textsuperscript{\#}\\
\citet{kabongo2024effective}&5.10&&0.23&&&&8.16&0.37&{TDMS-Ctx-v1}& TDMS-PR& Llama2 7B FS Full\textsuperscript{\#}\\
\citet{kabongo2024effective}&8.94&&6.36&&&&9.95&7.08&{TDMS-Ctx-v2}& TDMS-PR& Mistral 7B FS REC\textsuperscript{\#}\\
\citet{kabongo2024effective}&2.21&&1.87&&&&2.65&2.25&{TDMS-Ctx-v3}& TDMS-PR& Mistral 7B FS TAET\textsuperscript{\#}\\
\citet{kabongo2024effective}&9.6&&0.56&&&&14.52&0.84&{TDMS-Ctx-v1}& TDMS-PR& Mistral 7B FS Full\textsuperscript{\#}\\

\citet{Singh2024LEGOBENCH}&2.13&&&&&&&&{LEGOBench}&MS-PR\textsuperscript{\textdaggerdbl}& Mistral Instr. 7B\\
\citet{Singh2024LEGOBENCH}&1.81&&&&&&&&{LEGOBench}&MS-PR\textsuperscript{\textdaggerdbl}& Zephyr Beta 7B\\
\citet{Singh2024LEGOBENCH}&13.87&&&&&&&&{LEGOBench}&MS-PR\textsuperscript{\textdaggerdbl}& Gemini Pro \\
\citet{Singh2024LEGOBENCH}&13.06&&&&&&&&{LEGOBench}&MS-PR\textsuperscript{\textdaggerdbl}& GPT-4\\

\bottomrule
\multicolumn{12}{l}{\rule{0pt}{1.5ex} \textbf{Hybrid Domain Problem Framing}}\\[0.1ex]
\toprule

\citet{sahinuc2024efficient}&23.75&31.61&27.12&&&&&&SciLead& TDMR-PR & Llama 2\\ 
\citet{sahinuc2024efficient}&44.62&41.75&43.13&&&&&&SciLead& TDMR-PR & Mixtral \\ 
\citet{sahinuc2024efficient}&39.50&49.56&43.96&&&&&&SciLead& TDMR-PR & Llama 3 \\ 
\citet{sahinuc2024efficient}&70.34&68.22&69.26&&&&&&SciLead& TDMR-PR & GPT-4 \\ 

\bottomrule

\multicolumn{12}{p{16cm}}{\textsuperscript{\textdaggerdbl}Conditional on \(\langle\)task, dataset, metric\(\rangle\). \textsuperscript{\#} REC, TAET, and Full refer to DocREC, DocTAET, and the Full Paper representations of the document, respectively. These are reported as part of an ablation study examining different document representations. For more details on these representations, see \autoref{sec:doc_rep}.}

\end{tabular}
\caption{\fontsize{10pt}{12pt}\selectfont Summary of results for leaderboard \textbf{$\langle$Score$\rangle$} extraction. Notations: \textbf{FS} = Few Shot, \textbf{ZS} = Zero Shot, \textbf{Instr.} = Instruction.}
\end{table*}

\begin{table*}

\setlength{\tabcolsep}{1.0pt}
\fontsize{8pt}{10pt}\selectfont 

\renewcommand{\arraystretch}{0.9}
\begin{tabular}{l N N N N N N N N L{2.2cm} L{2cm} p{2.8cm}}
\toprule
&\multicolumn{3}{c}{\textbf{Micro}} & \multicolumn{3}{c}{\textbf{Macro}}& \multicolumn{2}{c}{\textbf{Part. Micro}} &&\\
\cmidrule(lr){2-4} \cmidrule(lr){5-7} \cmidrule(lr){8-9} 
\textbf{Reported In}& \textbf{P} &\textbf{R} & \textbf{F1} & \textbf{P} & \textbf{R} & \textbf{F1} & \textbf{P} & \textbf{F1} & \textbf{Dataset} & \textbf{Method}& \textbf{Experimental Setup}\\

\bottomrule

 \multicolumn{12}{l}{\rule{0pt}{1.5ex} \textbf{Open Domain Problem Framing}}\\[0.1ex]
\toprule

\citet{Singh2024LEGOBENCH}&&0.010&&&&&&&{LEGOBench}&MS-PR\textsuperscript{\textdaggerdbl}& Falcon 7B\\
\citet{Singh2024LEGOBENCH}&&0.002&&&&&&&{LEGOBench}&MS-PR\textsuperscript{\textdaggerdbl}& Falcon Instr. 7B\\
\citet{Singh2024LEGOBENCH}&&0.000&&&&&&&{LEGOBench}&MS-PR\textsuperscript{\textdaggerdbl}& Galactica 7B\\
\citet{Singh2024LEGOBENCH}&&0.024&&&&&&&{LEGOBench}&MS-PR\textsuperscript{\textdaggerdbl}& Llama 2 7B\\
\citet{Singh2024LEGOBENCH}&&0.077&&&&&&&{LEGOBench}&MS-PR\textsuperscript{\textdaggerdbl}& Llama 2 Chat 7B\\
\citet{Singh2024LEGOBENCH}&&0.351&&&&&&&{LEGOBench}&MS-PR\textsuperscript{\textdaggerdbl}& Mistral 7B\\
\citet{Singh2024LEGOBENCH}&5.75&20.42&&&&&&&{LEGOBench}&MS-PR\textsuperscript{\textdaggerdbl}& Mistral Instr. 7B\\
\citet{Singh2024LEGOBENCH}&&0.023&&&&&&&{LEGOBench}&MS-PR\textsuperscript{\textdaggerdbl}& Vicuna 7B\\
\citet{Singh2024LEGOBENCH}&1.49&10.87&&&&&&&{LEGOBench}&MS-PR\textsuperscript{\textdaggerdbl}& Zephyr Beta 7B\\
\citet{Singh2024LEGOBENCH}&&0.014&&&&&&&{LEGOBench}&MS-PR\textsuperscript{\textdaggerdbl}& Llama 2 13B\\
\citet{Singh2024LEGOBENCH}&&0.02&&&&&&&{LEGOBench}&MS-PR\textsuperscript{\textdaggerdbl}& Llama 2 Chat 13B\\
\citet{Singh2024LEGOBENCH}&&0.06&&&&&&&{LEGOBench}&MS-PR\textsuperscript{\textdaggerdbl}& Vicuna 13B\\
\citet{Singh2024LEGOBENCH}&2.73&3.38&&&&&&&{LEGOBench}&MS-PR\textsuperscript{\textdaggerdbl}& Gemini Pro\\
\citet{Singh2024LEGOBENCH}&17.14&25.24&&&&&&&{LEGOBench}&MS-PR\textsuperscript{\textdaggerdbl}& GPT-4\\
\bottomrule
\addlinespace[1ex]

\multicolumn{12}{p{16cm}}{\textsuperscript{\textdaggerdbl}Conditional on \(\langle\)task, dataset, metric\(\rangle\).}\\

\end{tabular}

 \caption{\fontsize{10pt}{12pt}\selectfont Summary of results for leaderboard \textbf{$\langle$Method$\rangle$} extraction.\label{tab:method}}

\end{table*}

\end{document}